\documentclass[sigconf, nonacm]{acmart}

\newcommand\vldbdoi{XX.XX/XXX.XX}
\newcommand\vldbpages{XXX-XXX}
\newcommand\vldbvolume{14}
\newcommand\vldbissue{1}
\newcommand\vldbyear{2026}
\newcommand\vldbauthors{\authors}
\newcommand\vldbtitle{\shorttitle} 
\newcommand\vldbavailabilityurl{}
\newcommand\vldbpagestyle{plain} 

\AtBeginDocument{%
  }

\usepackage[]{hyperref}
\usepackage{graphicx}
\usepackage{xcolor}
\usepackage{listings}
\usepackage{xspace}
\usepackage{amsmath,amsfonts, setspace,amsthm}
\usepackage{algpseudocode}
\usepackage{enumitem}
\usepackage{multirow}
\usepackage{tabularx, booktabs}
\usepackage{subcaption}
\usepackage{makecell}
\usepackage{wasysym}
\usepackage[scaled=0.85]{beramono}
\usepackage{graphicx}
\usepackage{xcolor}
\usepackage{listings}
\usepackage{xspace}
\usepackage{amsmath,amsfonts, setspace}
\usepackage{algpseudocode}
\usepackage{enumitem}
\usepackage{multirow}
\usepackage{tabularx, booktabs}
\usepackage{subcaption}
\usepackage{makecell}
\usepackage{algorithm}
\usepackage[scaled=0.85]{beramono}
\usepackage{wrapfig}
\usepackage{bm}
\usepackage{balance}

\newcommand{\stitle}[1]{{\noindent \bf {#1}}}

\definecolor{dkgreen}{rgb}{0,0.6,0}
\definecolor{gray}{rgb}{0.5,0.5,0.5}
\definecolor{mauve}{rgb}{0.58,0,0.82}
\definecolor{awesome}{rgb}{1.0, 0.13, 0.32}
\definecolor{beaver}{rgb}{0.62, 0.51, 0.44}
\definecolor{carrotorange}{rgb}{0.93, 0.57, 0.13}
\definecolor{chocolate}{rgb}{0.82, 0.41, 0.12}
\definecolor{copper}{rgb}{0.72, 0.45, 0.2}
\definecolor{crimsonglory}{rgb}{0.75, 0.0, 0.2}

\theoremstyle{definition}

\newtheorem{definition}{Definition}

\lstdefinestyle{base}{
  language=C,
  emptylines=1,
  breaklines=true,
  basicstyle=\ttfamily\color{blue},
  moredelim=**[is][\color{red}]{@}{@},
}

\newcommand{\projectname}{{GLM}\xspace}

\usepackage[normalem]{ulem}

\usepackage{titlesec}

\titlespacing*{\section}    {0pt}{.65\baselineskip plus .20\baselineskip minus .20\baselineskip}{.40\baselineskip}
\titlespacing*{\subsection} {0pt}{.55\baselineskip plus .15\baselineskip minus .15\baselineskip}{.35\baselineskip}
\titlespacing*{\subsubsection}{0pt}{.45\baselineskip plus .10\baselineskip minus .10\baselineskip}{.25\baselineskip}

\begin{document}

\title{Scaling Graph Chain-of-Thought Reasoning: A Multi-Agent Framework with Efficient LLM Serving}

\author{
Chengying Huan$^{1}$,
Ziheng Meng$^{1}$,
Yongchao Liu$^{2}$,
Zhengyi Yang$^{3}$,
Shipeng Li$^{1}$,
Xiabao Wu$^{2}$,
Haitao Zhang$^{2}$,
Yue Yun$^{2}$,
Yun Zhu$^{4}$,
Shaonan Ma$^{5}$,
Rong Gu$^{1}$,
Chuntao Hong$^{2}$,
Guihai Chen$^{1}$,
Chen Tian$^{1}$}

\affiliation{
$^{1}$Nanjing University;
$^{2}$Ant Group;
$^{3}$University of New South Wales;\\
$^{4}$Shanghai Artificial Intelligence Laboratory;
$^{5}$Tsinghua University
}

\begin{abstract}

Graph Chain-of-Thought (Graph-CoT) enables large language models (LLMs) to perform step-by-step reasoning over graph-structured knowledge, but existing pipelines suffer from low accuracy, excessive token usage, high latency, and low throughput due to single-agent monolithic prompts, repeated context re-encoding, and inefficient serving execution. 
We present \projectname, the first multi-agent Graph-CoT system co-designed with an optimized LLM serving architecture. \projectname decomposes reasoning into specialized agents for classification, reasoning, action generation, and graph retrieval, enabling branching and selective context sharing to reduce prompt length and reasoning iterations while preserving reasoning quality, thereby improving accuracy and reducing overall token consumption. 
To scale inference, we introduce a Graph-CoT–aware LLM inference mechanism with graph-specific KV-cache management, priority-based eviction, and pipelined execution to improve serving efficiency. 
Experiments demonstrate that \projectname improves answer accuracy by up to \textbf{38\%}, reduces token cost by up to \textbf{95.7\%}, lowers inference latency by \textbf{90.3\%}, and achieves up to \textbf{15.1$\times$} higher throughput compared to state-of-the-art Graph-CoT baselines, enabling efficient adoption for complex real-world reasoning at scale.

\end{abstract}

\maketitle

\pagestyle{\vldbpagestyle}
\begingroup\small\noindent\raggedright\textbf{PVLDB Reference Format:}\\
\vldbauthors. \vldbtitle. PVLDB, \vldbvolume(\vldbissue): \vldbpages, \vldbyear.\\
\href{https://doi.org/\vldbdoi}{doi:\vldbdoi}
\endgroup
\begingroup
\renewcommand\thefootnote{}\footnote{\noindent
This work is licensed under the Creative Commons BY-NC-ND 4.0 International License. Visit \url{https://creativecommons.org/licenses/by-nc-nd/4.0/} to view a copy of this license. For any use beyond those covered by this license, obtain permission by emailing \href{mailto:info@vldb.org}{info@vldb.org}. Copyright is held by the owner/author(s). Publication rights licensed to the VLDB Endowment. \\
\raggedright Proceedings of the VLDB Endowment, Vol. \vldbvolume, No. \vldbissue\ %
ISSN 2150-8097. \\
\href{https://doi.org/\vldbdoi}{doi:\vldbdoi} \\
}\addtocounter{footnote}{-1}\endgroup

\ifdefempty{\vldbavailabilityurl}{}{
\vspace{.3cm}
\begingroup\small\noindent\raggedright\textbf{PVLDB Artifact Availability:}\\
Code and data are available at \url{\vldbavailabilityurl}.
\endgroup
}

\section{Introduction}

\begin{figure}[t]
    \centerline{\includegraphics[width=1\linewidth]{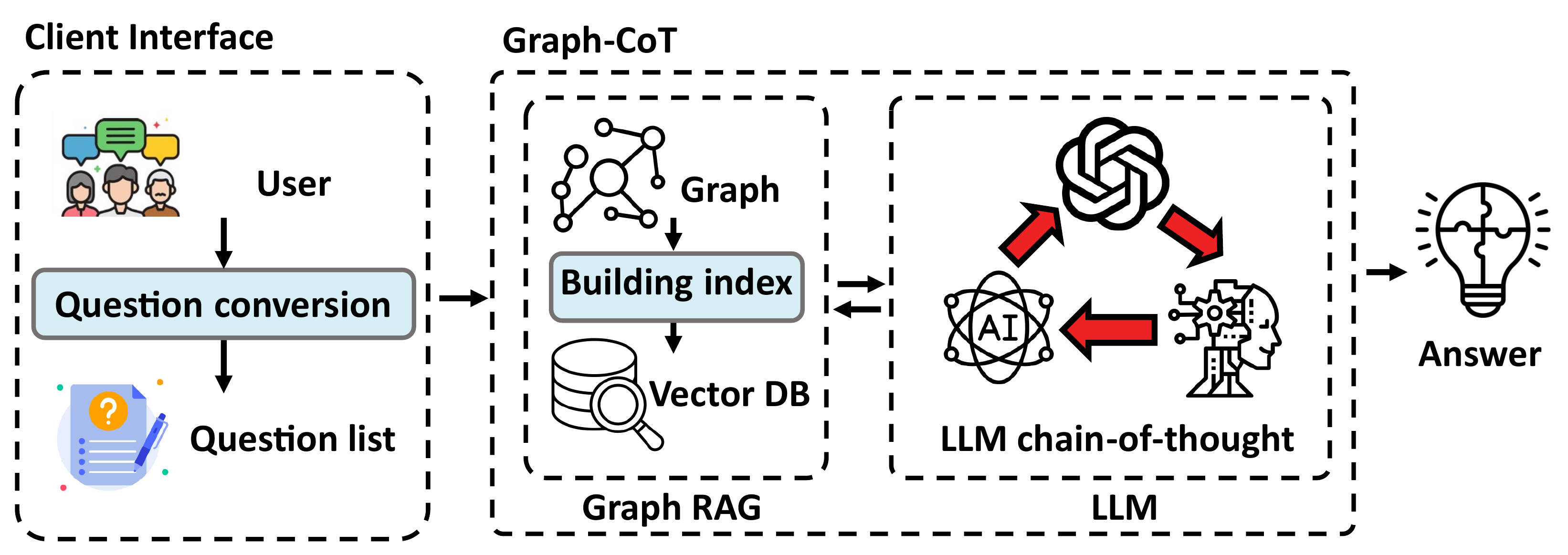}}
    \caption{Graph-CoT Framework.}
    \label{fig:introduction}
\end{figure}

Large Language Models (LLMs) such as GPT-4~\cite{achiam2023gpt}, Qwen~\cite{qwen}, Gemini~\cite{team2023gemini}, and DeepSeek~\cite{deepseekai2024deepseekv3technicalreport} have demonstrated remarkable general-purpose capabilities across multiple domains, including vibe coding~\cite{liu2023your,jiang2024self,jiang2024survey,zhang2023planning}, ChatBot~\cite{thoppilan2022lamda,hosseini2020simple,zhang2024comprehensive,yang2024zhongjing}, and complex reasoning~\cite{plaat2024reasoning,huang2022towards,lai2024lisa,lu2023chameleon}. 
In practice, reasoning ability and quality have become crucial determinants of the utility of LLMs for real-world applications such as program synthesis~\cite{gehring2024rlef,wang2023review}, robotic planning~\cite{zawalski2024robotic,cheng2024empowering}, and autonomous agents~\cite{deng2023mind2web,wang2024survey}.

To support complex reasoning, Chain-of-Thought (CoT) prompting~\cite{wei2022chain} has emerged as a core mechanism, enabling step-by-step inference by decomposing complex tasks into intermediate steps and thereby emulating human problem-solving processes.
Despite these advances, LLMs remain constrained by hallucination and limited access to domain-specific, proprietary, or real-time information. To mitigate these limitations, Retrieval-Augmented Generation (RAG)~\cite{lewis2020retrieval,guu2020retrieval,asai2024self} has become a widely adopted paradigm, enhancing factuality and grounding by integrating external retrieval into the generation process.

\stitle{Graph-CoT.}
Conventional RAG pipelines operate primarily on independent text chunks (i.e., flat text), overlooking the rich structural and relational dependencies across entities that are prevalent in real-world data such as knowledge graphs, enterprise data lakes, scientific repositories, financial networks, and other structured datasets. 
To bridge this gap, the state-of-the-art work on \textit{Graph Chain-of-Thought} (Graph-CoT) reasoning~\cite{jin2024graph} extends RAG by tightly integrating LLM reasoning with graph retrieval as illustrated in Figure \ref{fig:introduction}, enabling iterative interaction between LLMs and graph data. 
Specifically, Graph-CoT allows LLMs to iteratively query graph nodes, inspect attributes, explore neighbors, and accumulate evidence along graph structures. 
This paradigm supports multi-step relational reasoning beyond static text retrieval and highlights a promising direction for enabling data-driven LLM reasoning over complex, structured information sources~\cite{li2024graphreader,jin2024graph,sun2023think}.

\stitle{Limitations.} 
While Graph-CoT introduces a compelling paradigm for structured reasoning, current approaches face two fundamental limitations for effective application at scale:

\stitle{\underline{High Token Cost with Limited Accuracy Gains.}} 
State-of-the-art Graph-CoT frameworks that employ a \textit{single-agent} design which incur substantial token overhead yet yield limited accuracy gains. 
By consolidating all reasoning functions into a single prompt, these approaches maintain contextual coherence but are vulnerable to the ``lost-in-the-middle'' effect, where important information is overlooked as the context grows—particularly in Graph-CoT, where the prompt contains both long natural-language context and retrieved graph information. 
This often results in unstable reasoning and incorrect decisions, especially on complex multi-hop tasks. 
To maintain accuracy, existing systems rely on multiple rounds of iterative reasoning; however, each step re-encodes the full context and intermediate traces, resulting in substantial token overhead. 
Achieving moderate accuracy often requires long reasoning chains and repeated prompt extensions, frequently exceeding 40{,}000 tokens per query. With commercial models such as GPT-4.5~\cite{openai_pricing}, this can cost over \$3 per query, making single-agent Graph-CoT economically infeasible at scale. 
Despite this high token budget, accuracy improvements remain modest, highlighting the inefficiency of single-agent CoT for structured graph reasoning. 
On GRBench~\cite{jin2024graph}, existing Graph-CoT systems achieve R–L values (the proportion of correct reasoning steps relative to total chain length) between 31\% and 46\% (Table~\ref{tab:cot-performance}), falling short of the commonly accepted 50\% threshold for reliable LLM reasoning~\cite{chen2024benchmarking,vendrow2025largelanguagemodelbenchmarks}. 
This indicates that simply increasing token consumption does not reliably yield better graph reasoning performance.

\stitle{\underline{High Latency with Poor Throughput Scalability.}} 
The Graph-CoT framework suffer from significant efficiency bottlenecks. In our preliminary experiments, we observe that current Graph-CoT methods exhibit end-to-end latency ranging from 11s to 39s per query (Table~\ref{tab:cot-performance}), which is far below what is acceptable for real-time or interactive applications such as question answering or conversational agents, where low-latency, second-scale responsiveness is typically required. This inefficiency mainly arises from the combined cost of large \textit{LLM inference overhead} and the long \textit{graph retrieval latency} inherent in the Graph-CoT workflow. 
 Existing Graph-CoT frameworks do not effectively leverage KV caching, despite its proven benefits in modern LLM serving. Directly applying standard cache mechanisms, such as those in vLLM~\cite{kwon2023efficient}, is ill-suited for Graph-CoT, as each reasoning step is prompted and re-encoded independently, preventing the reuse of previously computed hidden states and forcing redundant decoding over similar contexts. 
Furthermore, Graph-CoT reasoning unfolds step-by-step, producing long and highly variable intermediate traces that disrupt prefix sharing, leading to low cache reuse, substantial computational redundancy, and, ultimately, degraded throughput. 
In addition, graph knowledge is retrieved at fine granularity (e.g., per-node or per-edge), yielding fragmented content that rarely aligns into common prefixes across steps or queries, resulting in low KV-cache reuse even when multiple queries contain semantically similar facts. 
Finally, graph retrieval incurs a high cost (up to 23\% of the total time as shown in Figure~\ref{fig:end_to_end_latency_comparison}) and must be executed across multiple sequential rounds. The combination not only increases end-to-end latency but also severely limits throughput, preventing efficient execution at scale.

\begin{table}[t]
\centering
\caption{Graph-CoT and GLM Performance Comparison}\label{tab:cot-performance}
\label{tab:cot-vs-glm}
\setlength{\tabcolsep}{4pt}
\renewcommand{\arraystretch}{0.9}
\small
\resizebox{\linewidth}{!}{%
\begin{tabular}{@{}lccc ccc@{}}
\toprule
 & \multicolumn{3}{c}{\textbf{Graph-CoT}} & \multicolumn{3}{c}{\textbf{GLM (Ours)}} \\
\cmidrule(lr){2-4}\cmidrule(lr){5-7}
Domain & Cost (\$) & Acc. (R-L) & Latency (s) & Cost (\$) & Acc. (R-L) & Latency (s) \\
\midrule
Academic    & 1.9 & 0.31 & 27.2 & 0.1 & 0.55 & 3.0 \\
E-commerce  & 1.9 & 0.39 & 17.8 & 0.1 & 0.77 & 3.1 \\
Literature  & 1.7 & 0.42 & 11.3 & 0.1 & 0.65 & 2.8 \\
Healthcare  & 3.6 & 0.33 & 38.6 & 0.2 & 0.62 & 3.4 \\
Legal       & 2.8 & 0.46 & 22.8 & 0.2 & 0.63 & 5.9 \\
\bottomrule
\end{tabular}}
\end{table}

\stitle{Our Solution and Contributions.} 
To address the limitations, we revisit existing Graph-CoT designs and jointly co-design the reasoning framework and LLM serving architecture, incorporating system-level optimizations to enable scalable Graph-CoT reasoning.
We present \textbf{\projectname}, a \textit{multi-agent} \underline{G}raph-CoT \underline{L}L\underline{M} framework with an optimized LLM serving that improves accuracy, latency, and token efficiency. 
\projectname extends Graph-CoT with a multi-agent design and integrates code generation and selective branch execution to reduce token usage and end-to-end latency while improving answer accuracy. 
It organizes reasoning as a directed graph of fine-grained, interdependent steps executed by specialized agents, generalizing linear chain-of-thought into a more expressive Graph-CoT abstraction. 
We further introduce Graph-CoT–specific KV-cache management and pipelined execution to enhance runtime efficiency, significantly reducing latency and increasing throughput. 
Together, these techniques improve accuracy, reduce token consumption, and accelerate inference. Experimental results demonstrate that this co-designed reasoning-and-serving architecture enables LLMs to scale to complex graph reasoning tasks both effectively and efficiently.
The key contributions of this paper are as follows:


\begin{itemize}[leftmargin=*, wide, itemindent=0pt, nosep]


    \item \textbf{Multi-Agent Graph-CoT Framework.} 
    We propose \projectname, the \textit{first} multi-agent Graph-CoT reasoning framework. 
    \projectname employs four collaborative agents, including a Graph RAG retriever and three LLM-based agents: a classification agent, a reasoning agent, and an action agent. 
    This design enables branching and selective context sharing, reduces the number of reasoning iterations, and substantially shortens prompts while preserving reasoning quality, thereby improving LLM reasoning accuracy and reducing overall token consumption.

    \item \textbf{Graph-CoT–Aware LLM Inference.}
    We introduce a Graph-CoT–aware LLM inference mechanism that combines specialized KV-cache management with pipelined execution. 
    Specifically, we design a \textit{vertex-centric KV-cache reuse model} that caches coarse-grained vertex representations to maximize reuse across multi-agent reasoning, together with a \textit{priority-based KV-cache eviction policy} tailored to Graph-CoT access patterns to improve cache efficiency and reduce inference latency. 
    Additionally, we develop a \textit{pipelined execution strategy} that overlaps graph retrieval with LLM inference and enables parallel execution for concurrent queries, significantly improving end-to-end throughput.

    \item \textbf{Extensive Experimental Evaluation.} We conduct comprehensive experiments against four baseline methods on GRBench~\cite{jin2024graph}, a diverse graph reasoning benchmark covering five domains: academia, e-commerce, literature, healthcare, and law. Experimental results show that \projectname improves reasoning accuracy by up to \textbf{60\%}, \textbf{62\%}, \textbf{55\%} and \textbf{38\%} compared to Base LLM~\cite{qwen}, Text RAG~\cite{gao2023retrieval}, Graph RAG~\cite{ye2023language} and Graph-CoT~\cite{jin2024graph}, respectively. Furthermore, it substantially outperforms Graph-CoT across multiple key metrics: reducing token cost by up to \textbf{95.7\%}, lowering inference latency by \textbf{90.3\%}, and achieving up to \textbf{15.1$\times$} higher throughput.
\end{itemize}


\section{Background and Related Works}


\subsection{Problem Definition}

With the proliferation of large-scale graph data across diverse real-world applications, an increasing number of studies leverage graphs to enhance reasoning capabilities in LLMs \cite{wu2024can, cao2024graphreason, jin2024graph}.
A graph reasoning framework enables LLMs to iteratively reason over graph-structured data, allowing them to access domain-specific knowledge and improve answer accuracy.

\label{sec:problem-definition}
\stitle{Graph Reasoning Problem.} 
Let $q \sim \mathcal{D}$ denote a query drawn from a distribution $\mathcal{D}$, and let $G$ be an external graph that can be accessed during reasoning. 
We consider a set of agents $\mathcal{A}=\{a_1,\dots,a_K\}$ coordinated by a policy $\pi$. 
Given a query $q$, the interaction among the policy $\pi$, the agents $\mathcal{A}$, and the graph $G$ induces a stochastic reasoning trajectory
$\tau \sim P^{\pi,\mathcal{A}}(\,\cdot \mid q, G)$.
An answer $\hat{y}$ is generated from the trajectory through a decoding function $g$, i.e., $\hat{y}=g(\tau)$.
Correctness is evaluated by an accuracy function $Acc(\hat{y}, y^\star(q)) \in [0,1]$, where $y^\star(q)$ denotes the ground-truth answer.
Efficiency is measured by the total token cost $Tok(\tau)$ and end-to-end latency $Lat(\tau)$.

The optimization objective is to design a policy $\pi$ and agent set $\mathcal{A}$ that maximize expected accuracy under bounded cost and latency constraints:
\begin{equation}
\begin{array}{cl}
\max_{\pi,\,\mathcal{A}} 
& \mathbb{E}_{q \sim \mathcal{D}}\, \mathbb{E}_{\tau \sim P^{\pi,\mathcal{A}}(\,\cdot \mid q, G)} \big[ Acc(g(\tau), y^{\star}(q)) \big] \\[3pt]
\text{s.t.} 
& \mathbb{E}_{q, \tau}[Tok(\tau)] \leq \tau_{\max}, \\[2pt]
& \mathbb{E}_{q, \tau}[Lat(\tau)] \leq \ell_{\max}.
\end{array}
\end{equation}

In practice, a trajectory $\tau=(s_0, u_0, o_1, \dots, s_T)$ records the sequence of states $s_t$, actions $u_t$ (e.g., selecting an agent or issuing a graph operation), and observations $o_{t+1}$ returned by either an agent $a_k \in \mathcal{A}$ or the graph $G$. 
The policy $\pi$ governs routing decisions, agent selection, memory management, and termination.
The total token cost and latency are instantiated as:
$
Tok(\tau)=\sum_{t=0}^{T}\!\big(\mathrm{tok\_in}(u_t)+\mathrm{tok\_out}(o_{t+1})\big), 
Lat(\tau)=\sum_{t=0}^{T}\!\Delta t_t.
$

\noindent\textit{Remark.} Unlike most prior work, which pays limited attention to token cost and efficiency (i.e., end-to-end latency and throughput), we explicitly model both via $\mathrm{Tok}(\tau)$ and $\mathrm{Lat}(\tau)$ and treat them as top-priority optimization objectives in our design.

\subsection{Existing Graph Reasoning Methods} 

Recent research has explored how to design effective policies and agent architectures to improve reasoning accuracy over graph-structured data~\cite{li2024graphreader, wang2023knowledgpt, jiang2023structgpt, jin2024graph}. 
\textsc{GraphReader}~\cite{li2024graphreader} adopts a planning-based, coarse-to-fine exploration strategy that leverages predefined functions to read node contents and neighbourhoods, while maintaining iterative note-taking and reflection. 
\textsc{KnowledGPT}~\cite{wang2023knowledgpt} formulates graphs and queries as executable programs, using a Program-of-Thought framework to generate and execute search code that interacts with a lightweight knowledge-base API for multi-hop reasoning. 
\textsc{StructGPT}~\cite{jiang2023structgpt} employs an Iterative Reading–then–Reasoning (IRR) loop, in which task-specific interfaces extract graph evidence that the LLM subsequently reasons over after linearization. 
However, these approaches primarily emphasize reasoning accuracy, often overlooking efficiency metrics such as token consumption and end-to-end latency.

Among existing methods, \textsc{Graph-CoT}~\cite{jin2024graph} represents the current state-of-the-art solution for graph reasoning. 
It instantiates an iterative \emph{Reasoning–Interaction–Execution} framework, enabling the LLM to autonomously determine subsequent graph operations, issue primitive function calls, execute them on an external graph, and repeat this process until convergence. 
This explicit stepwise traversal allows the model to exploit graph topology dynamically, contrasting with static retrieval-augmented generation (RAG) approaches that rely on context stuffing without structural awareness.

\subsection{RAG and CoT}

\stitle{Retrieval-Augmented Generation (RAG).}  
RAG enhances the response quality of large language models (LLMs) by integrating external knowledge sources during inference, which is particularly beneficial for knowledge-intensive tasks~\cite{huang2025survey}. 
A typical RAG pipeline consists of two stages: \textit{retrieval} and \textit{generation}~\cite{gao2023retrieval}. 
In the retrieval stage, external documents or knowledge bases are first encoded into vector representations and indexed for efficient similarity search~\cite{10.5555/3495724.3496517}. 
Given a query $Q$, the most relevant entities are retrieved and concatenated with $Q$ to form an augmented prompt. 
This enriched prompt is then passed to the LLM, enabling it to generate responses that are more contextually grounded.

\stitle{Chain-of-Thought (CoT).}  
CoT prompting~\cite{wei2022chain} is a widely adopted technique for improving the reasoning capability of LLMs~\cite{chen2025towards}, particularly in complex or multi-step reasoning tasks. 
Instead of directly producing an answer, CoT encourages the model to generate intermediate reasoning steps in a structured format (e.g., \textit{<input, thoughts, output>}). 
These intermediate steps guide the model through reasoning processes such as arithmetic computation, commonsense inference, and symbolic deduction~\cite{wang2022self, lyu2023faithful}. 
Compared to conventional few-shot prompting, CoT enhances both interpretability and performance by decomposing complex reasoning into a sequence of transparent and verifiable sub-tasks.

\subsection{LLM Inference and Prefix Caching}\label{sec:kv_cache_background}

During LLM inference, computation proceeds in two stages: the \textbf{Prefill} stage and the \textbf{Decoding} stage. 
In the prefill stage, the entire input sequence $X = [x_1, x_2, \ldots, x_s]$ is processed to initialize the key–value (KV) cache. 
At each transformer layer, token embeddings are projected into query ($Q$), key ($K$), and value ($V$) representations. 
The $K$ and $V$ tensors are stored in the KV cache for efficient reuse during subsequent decoding. 
Once the first output token $x_{s+1}$ is produced, the model transitions into the autoregressive decoding phase, generating tokens sequentially based on previously generated ones while incrementally updating the KV cache.

Prefix caching is a widely adopted optimization technique for reducing redundant computation in LLM inference. 
The key idea is to cache the KV-cache blocks corresponding to previously processed input prefixes and reuse these cached blocks when new requests share identical prefixes, therefore reducing latency and GPU usage while preserving outputs. 
When GPU memory is limited, cached blocks are evicted according to predefined policies, with Least Recently Used (LRU) being the most common strategy.

\section{Motivation and Key Insights}

\begin{figure}[t]
    \centerline{\includegraphics[width=1\linewidth]{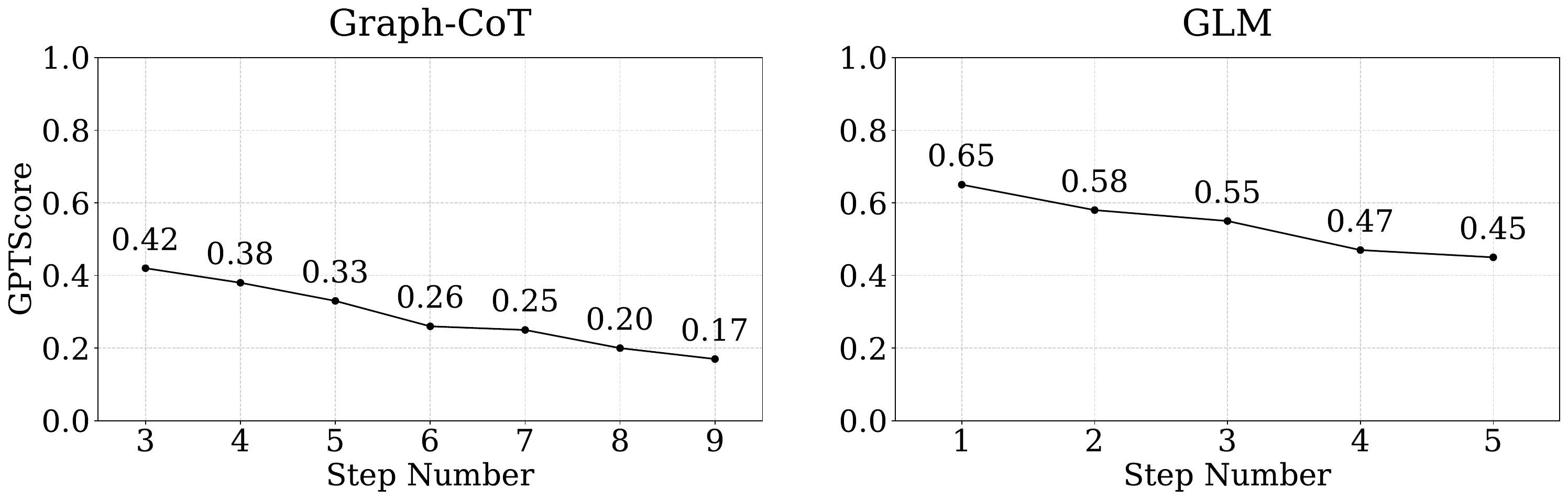}}
    \caption{Sensitivity of accuracy on reason steps.}
    \label{fig:step_accuracy}
\end{figure}

In this section, we first conclude the \textit{limitations of existing single-agent Graph-CoT frameworks} and then summarize the \textit{key insights that motivate our proposed system, \projectname}.

\textbf{Limitations of Graph-CoT.}  
Current frameworks struggle with queries requiring multi-hop reasoning for two reasons: (i) the limitations of the single-agent paradigm and (ii) inefficiencies in the LLM inference stack.  
(1) \textbf{Single-agent limitations.} Existing Graph-CoT frameworks struggle with complex multi-hop reasoning under a single-agent architecture. As reasoning steps increase, input sequences grow longer and accumulate redundant context, causing information dilution and the “lost-in-the-middle” problem. Consequently, Graph-CoT accuracy drops sharply with increasing step depth, e.g., the GPTScore~\cite{fu2023gptscore} declines from 0.42 at 3 steps to 0.17 at 9 steps (Figure~\ref{fig:step_accuracy}). In contrast, our multi-agent design achieves comparable reasoning with fewer steps and greater accuracy. Moreover, existing Graph-CoT frameworks suffer from high token overhead due to repeated prefixes and growing context (e.g., \$1.9$\sim$\$3.6 per GPT-4.5 query; see Table~\ref{tab:cot-performance}).
(2) \textbf{Inference inefficiencies.} Under concurrency, KV-cache hit rates are modest (e.g., 61\% with \textsc{vLLM} in our measurements), and naive LRU eviction, along with non-canonical ordering of entity facts, impedes prefix reuse; 
retriever latency also increases with graph size. As a result, Graph-CoT exhibits end-to-end latencies of up to 39\,s per query (Table~\ref{tab:cot-performance}), which is far from sufficient for real-time or interactive applications.

Based on the limitations and observations of existing approaches, we derive the following key insights and core design principles:



\textbf{Key Insights.}
Based on the above limitations, we make two key insights: 
(i) a \emph{multi-agent framework} that separates retrieval from reasoning, routes deterministic queries to skip steps, and replaces long CoT chains with single-pass code, thereby reducing redundant context and token cost while improving accuracy; and 
(ii) a \emph{co-designed LLM optimization} stack that maximizes KV-prefix reuse (via coarse node-fact chunks and reuse-aware priority eviction) and pipelines retrieval with decoding to hide retrieval latency, thereby reducing end-to-end latency and increasing throughput.

\textbf{Key Objective.}
Guided by the key insights, we build \projectname to make Graph-CoT both accurate and efficient. \projectname provides a multi-agent planning and runtime substrate for Graph-CoT (Section~\ref{sec:framework}) and a Graph-CoT–aware inference stack (Section~\ref{sec: optimization}). We specify the plan abstraction and cost model, cache and scheduling policies, 
a pipelined execution engine, and evaluate it on large graphs to demonstrate higher accuracy, lower token usage, lower end-to-end latency, and higher throughput than prior Graph-CoT systems.

\section{Multi-Agent Graph-CoT Framework}\label{sec:framework}

\begin{figure}[t]
    \centerline{\includegraphics[width=.8\linewidth]{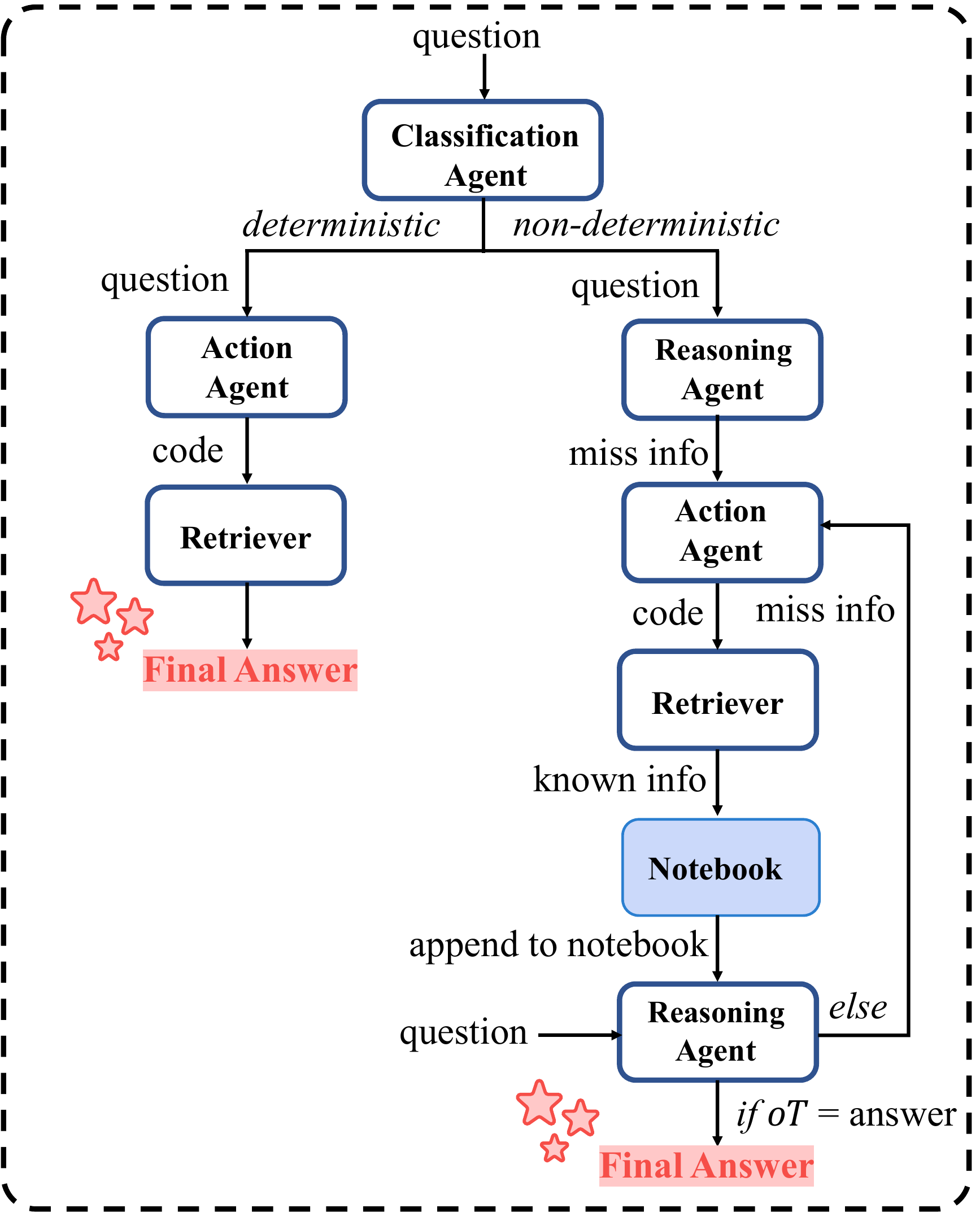}}
    \caption{Multi-agent reasoning framework workflow.}
    \label{fig:system_overview}
\end{figure}

\subsection{Overview}
We first introduce \projectname's multi-agent Graph-CoT framework, designed to improve accuracy while reducing token usage (Figure~\ref{fig:system_overview}). It comprises a Graph RAG retriever and three LLM agents: a classifier (C-Agent), a reasoner (R-Agent), and an action generator (A-Agent). Agents operate on task-specific inputs and invoke the inference engine to produce intermediate or final outputs, while the retriever supplies the necessary graph facts at each step. We next formalize the LLM-agent abstraction.


\begin{definition}[\textbf{LLM-based Agent}]
Given an LLM service defined as: $LLM: \text{Prompt} \rightarrow \text{Output}$, an LLM-based agent is a 3-tuple, $\text{X-Agent} = (I_x, P_x, O_x)$, where $I_x$ is the input space (e.g., outputs from upstream agents, user queries, or graph information), $P_x$ is the prompt template that maps $I_x$ to LLM instructions (enabling tasks such as reasoning, summarization, or code generation), and $O_x$ is the output space comprising all possible outputs of LLM. The execution mechanism of an agent involves applying its agent-specific prompt template \( P_x \) to a given input \( i \in I_x \), generating a formatted prompt that is then passed to the LLM to produce the final output $ o \in O_x $. Formally, this can be expressed as: $o = \text{LLM}(P_x(i)).$
\end{definition}

\begin{figure}[t]
    \centerline{\includegraphics[width=1\linewidth]{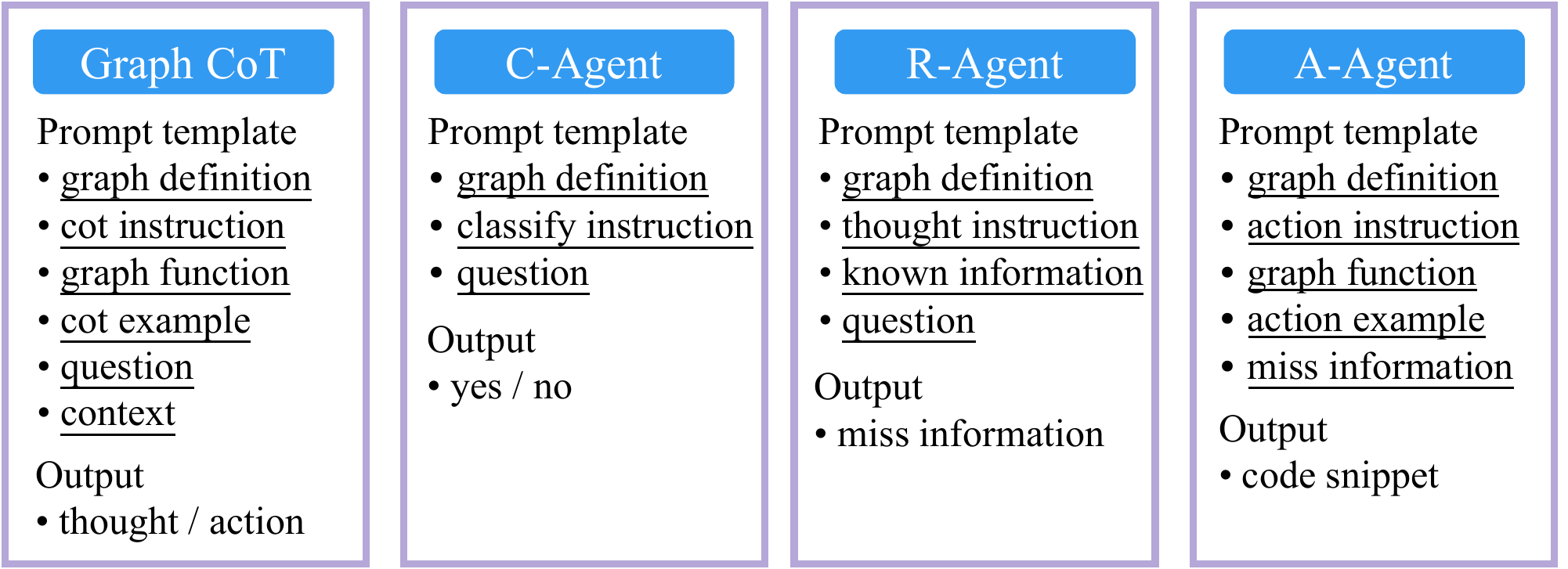}}
    \caption{Graph-CoT and GLM's agents comparison.}
    \label{fig:agent_prompt}
\end{figure}

Building on the agent definition above, we now introduce the specific agents used in our multi-agent reasoning framework, as illustrated in Figure~\ref{fig:agent_prompt}.

\textbf{Classification Agent (C-Agent).}
The \textit{Classification Agent} determines whether a question is \textit{deterministic} or \textit{non-deterministic}. Deterministic questions can be answered directly by retrieving information from the graph, such as “What is the price of A?”. Non-deterministic questions, like “Recommend the next item based on user history: A, B,” require multi-hop reasoning over multiple nodes and relationships.  

Formally, \(\text{C-Agent}=(I_c,P_c,O_c)\), where \(I_c\) is the natural-language question, \(P_c\) is the prompt template for classification, and \(O_c\in\{\text{deterministic},\text{non-deterministic}\}\).


\textbf{Reasoning Agent (R-Agent).}
The \textit{Reasoning Agent} is responsible for determining whether the currently known information is sufficient to answer a given question. If the available information is insufficient, the agent identifies what additional information is required. Otherwise, it proceeds to provide the final answer directly. The \textit{Reasoning Agent} operates through a \textit{notebook} mechanism, which accumulates and maintains known facts during the chain-of-thought reasoning process. At each step, the notebook is updated with new facts retrieved from the graph by the retriever.

Formally, the Reasoning Agent is defined as:
$
\text{R-agent} = (I_t, P_t, O_t),
$
where $I_t$ is the input space, consisting of the current notebook (i.e., known facts) and the user question; \( P_t \) is the prompt template, which encodes this information to guide reasoning; and \( O_t \) is the output space, representing either the final answer or a specification of additional information needed.

\textbf{Action Agent (A-Agent).}
The \textit{Action Agent} generates executable Python code snippets to retrieve the missing information identified by the \textit{Reasoning Agent}. Unlike prior Graph-CoT systems~\cite{jin2024graph}, which limit actions to predefined operations, our agent produces expressive code that can combine multiple functions, use basic control structures (e.g., \texttt{if-else}, \texttt{for}), and employ standard data types (\texttt{set}, \texttt{list}, \texttt{dict}). This allows complex reasoning to be completed in a single execution step, reducing multi-round interactions. The agent outputs only the essential results via \texttt{print()}, avoiding redundant intermediate context.

Formally, the \textit{Action Agent} is defined as:
$
\text{A-Agent} = (I_a, P_a, O_a),
$
where $I_a$ is the input space, representing the missing information specified by the \textit{Reasoning Agent}; \( P_a \) is the prompt template, which encodes the task of generating the corresponding code snippet; and \( O_a \) is the output space, consisting of the generated Python code.

\begin{algorithm} [t]
\caption{Multi-Agent Reasoning Workflow.}
\label{alg:workflow}
\small
\begin{algorithmic}[1]
\Require Question $q$, $\text{C-Agent}$, $\text{R-agent}$, $\text{A-Agent}$, Retriever, LLM service.
\State $o_c \gets LLM(P_c(q))$
\If{$o_c ==$ \text{yes}}
    \State $o_a \gets LLM(P_a(q))$
    \State $a \gets Retriever(o_a)$
\Else
    \State Initialize notebook $n \gets \{\}$
    \While{$True$}
        \State $o_t \gets LLM(P_t(n, q))$
        \If{$o_t$ indicates answer completeness}
            \State $a \gets o_t$
            \State \textbf{break}
        \Else
            \State $o_a \gets LLM(P_a(o_t)))$
            \State $o_r \gets Retriever(o_a)$
            \State $n \gets n \cup \{o_r\}$
        \EndIf
    \EndWhile
\EndIf
\State \Return final answer $a$
\end{algorithmic}
\end{algorithm}

\textbf{Graph RAG Retriever.}
In addition to the LLM-based agents, a key component of \projectname is the \textit{Graph RAG Retriever}, which bridges the \textit{Action Agent} and the underlying graph. When the \textit{Action Agent} generates a Python snippet to obtain missing information, it is executed via Python’s \texttt{exec()} function, and the results are appended to the agent’s notebook as new knowledge for subsequent reasoning. We build upon the Graph-CoT~\cite{jin2024graph} retrieval interface and extend it with a new function, \texttt{NodeInfo()}, which provides vertex-centric contextual information (see Section~\ref{sec:vertex_centric_kv}). Specifically, \texttt{RetrieveNode()} maps entities to node IDs through vector search over the graph’s embedding index (GPU-accelerated), while other functions map node IDs to their attributes and metadata stored in an in-memory dictionary. Table~\ref{tab:function_summary} summarizes all core functions, including our extensions.

\begin{table*}[ht]
\centering
\small
\begin{tabular}{|l|l|l|l|}
\hline
\textbf{Function} & \textbf{Input} & \textbf{Output} & \textbf{Function Description} \\
\hline
RetrieveNode & text & nodeID & Returns the ID of the node most relevant to the input text using vector similarity. \\
\hline
NodeInfo & nodeIDs & vertex chunk & Returns the vertex chunk for the given node. \\
\hline
NodeFeature & nodeIDs, featureName & values & Retrieves feature values for the given nodes. \\
\hline
NodeDegree & nodeID, neighbourType & degree & Returns the number of neighbours of the specified type for the given node. \\
\hline
neighbourCheck & nodeID, neighbourType & nodeIDs & Returns the neighbours of the specified type for the given node. \\
\hline
\end{tabular}
\caption{Summary of core graph-related functions in Graph RAG retriever.}
\label{tab:function_summary}
\end{table*}

\subsection{Workflow}
Upon receiving a question $q$, \projectname initiates a multi-agent reasoning workflow, coordinating interactions between LLM-based agents and the Graph RAG retriever, as detailed below:

\noindent \textbf{Step 1:} The \textit{Classification Agent} determines whether $q$ is deterministic or non-deterministic. If $q$ is deterministic, we proceed directly to Step~3. Otherwise, we move to Step~2.

\noindent \textbf{Step 2:} For non-deterministic questions, the \textit{Action Agent} generates a code snippet $o_a$ based on $q$. This snippet is executed by the \textit{Graph RAG retriever} to obtain relevant information from the graph dataset. The final answer is then extracted and returned using the \texttt{print} statement within the executed code.

\noindent \textbf{Step 3:} Initialize an empty notebook $n = \{\}$. The \textit{Reasoning Agent} takes $q$ and $n$ as input and produces the missing information $o_t$.

\noindent \textbf{Step 4:} Enable iterative chain-of-thought reasoning: (1) \textit{Action Agent:} Generate a code snippet $o_a$ from $o_t$; (2) \textit{Graph RAG Retriever:} Execute $o_a$ and append the result to the notebook $n$; (3) \textit{Reasoning Agent:} Generate the next $o_t$ based on $q$ and $n$.


This iterative process continues until a termination condition is met, whereby the output of \textit{Reasoning Agent} $o_t$ represents a complete answer to the original query. At that point, $o_t$ is returned as the final output of the system. The overall workflow is detailed in Algorithm~\ref{alg:workflow}.

\begin{figure}[t]
    \centerline{\includegraphics[width=1\linewidth]{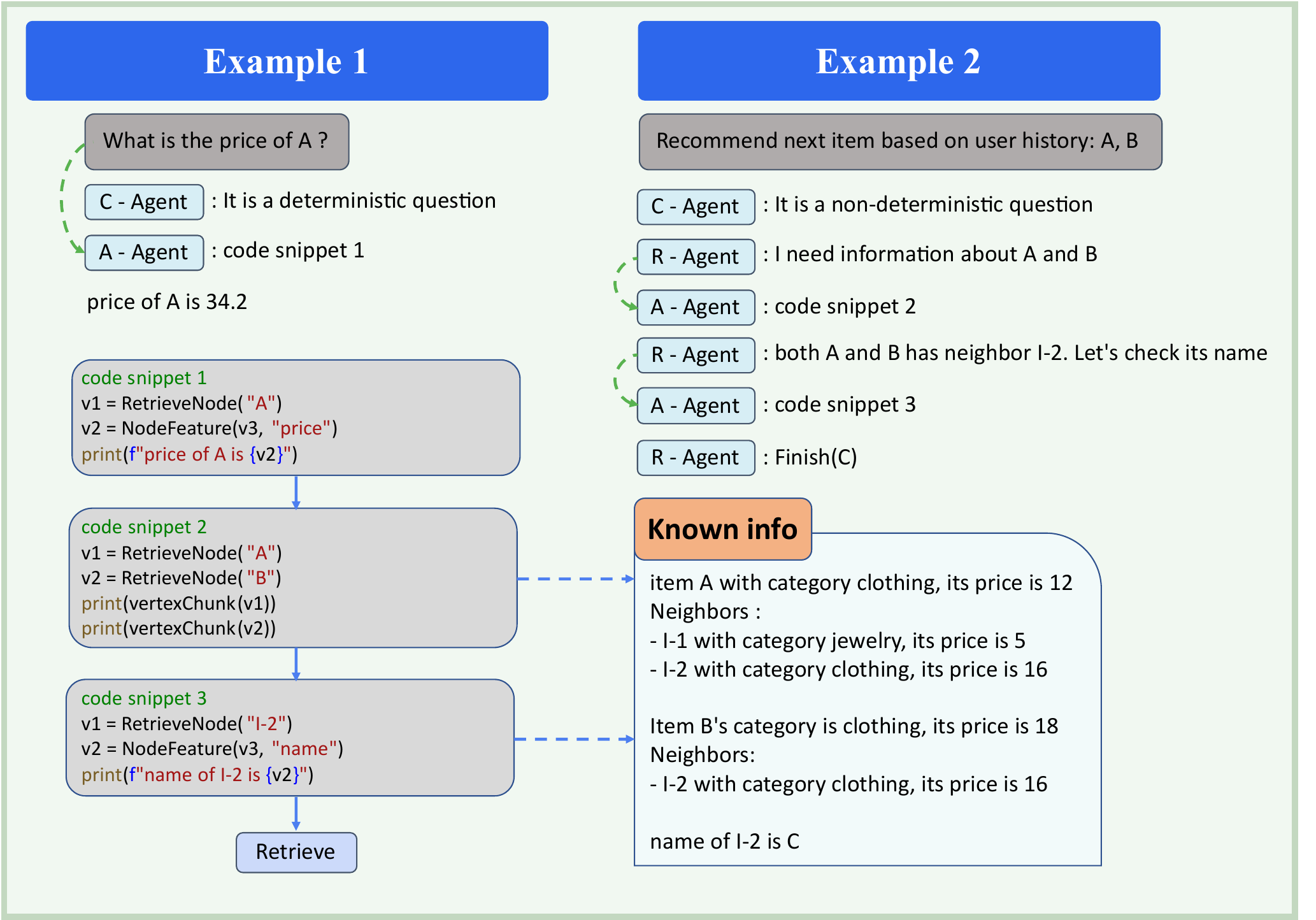}}
    \caption{Multi-agent reasoning example.}
    \label{fig:example}
\end{figure}

\textbf{Case Study.}
Figure~\ref{fig:example} shows two flows.
(1) \emph{Deterministic} (“What is the price of A?”): The C-Agent classifies the query as deterministic. The A-Agent produces a one-shot snippet that fetches and prints A’s price. The Retriever executes the snippet and returns the answer.
(2) \emph{Non-deterministic} (“Recommend next item given A,B”): The C-Agent classifies the query as non-deterministic. The R-agent first infers what additional information is required to answer the query, reasoning that full representations of items \textit{A} and \textit{B} are needed. The A-Agent generates a code snippet to retrieve this information, which is executed by the Retriever and appended to the known context. With the updated context, the R-agent observes that both \textit{A} and \textit{B} share item \textit{C}, and its category and price are also similar to those of the items the user has viewed. The R-agent concludes that \textit{Item C} is the most appropriate recommendation for the user.

\textbf{Cost Model - Token Usage Analysis.} 
Let \( T_{\text{G}} \), \( T_{\text{c}} \), \( T_{\text{a}} \), and \( T_{\text{t}} \) denote the average token consumption per invocation by the Graph-CoT agent, classification agent, action agent, and reasoning agent, respectively. We assume that \( T_{\text{c}} \ll T_{\text{G}} \). Additionally, as shown in Figure~\ref{fig:agent_prompt}, we approximate that \( T_{\text{a}} + T_{\text{t}} \approx T_{\text{G}} \).

For Graph-CoT, each reasoning step requires two invocations (thought and action), assume that for each question, Graph-CoT requires $k_1$ step, yielding a total of \( 2k_1 \cdot T_{\text{G}} \) tokens.

For \projectname, the token cost depends on the query type. For deterministic queries, only one classification and one action invocation are needed, i.e., \( T_{\text{c}} + T_{\text{a}} \). For non-deterministic queries, the classification agent is called once, while the action agent and reasoning agent are called \( k_2 \) and \( k_2 + 1 \) times, respectively. Based on the assumption, the total becomes approximately \( (k_2 + 1) \cdot T_{\text{G}} \).

As a result, the reduction compared to the baseline Graph-CoT is as follows:  
For deterministic queries, the reduction is approximately \( (2k_1 - 1) \cdot T_{\text{G}} \);  
For non-deterministic queries, the reduction is approximately \( (2k_1 - k_2 - 1) \cdot T_{\text{G}} \).  
These results further highlight the significant token-efficiency advantages of \projectname, demonstrating its superiority across both simple and complex reasoning tasks.

\subsection{Code Generation Optimization} 

Given a graph $g$ and a specification of \emph{missing information} $m$ produced by the Reasoning Agent, \textbf{code generation} defines a mapping $\mathcal{G}: (q, m) \rightarrow s$ that yields an \emph{executable} code snippet $s$.  

\textbf{Code Generation within Graph-CoT.}  
In Graph-CoT, code refers to a predefined set of graph-retrieval functions (\texttt{RetrieveNode}, \texttt{NodeFeature}, \texttt{neighbourCheck}, \texttt{NodeInfo}) that the LLM can invoke to obtain the required information from the graph.  

\textbf{Limitations of Existing Graph-CoT Code Generation.}  
In the original Graph-CoT framework, the LLM interacts with the external graph by selecting from a fixed set of predefined retrieval functions and providing their parameters. This design constrains retrieval granularity and leads to inefficiencies in two common cases. First, when a query requires information from multiple vertices, multiple reasoning rounds must be executed to collect all necessary data. Second, when the target information depends on intermediate computation (e.g., the intersection of two vertices’ neighbour lists), the system must explicitly retrieve and store all intermediate results, increasing both context size and token cost.  

\textbf{Our Approach.} 
To overcome these limitations, we introduce a code generation optimization that enables the Action Agent to synthesize complete executable Python snippets rather than merely selecting from predefined functions. Each snippet is composed only of: (i) the predefined graph-retrieval functions and basic control; (ii) optional \emph{local} computations (e.g., aggregation, set intersection) to derive the precise information required by $m$; and (iii) a minimal \texttt{print()} statement that outputs only the final result, avoiding unnecessary intermediate text. Executing $s$ once against the Graph RAG retriever produces the required facts to update the agent’s notebook or generate the final answer, thereby replacing multiple reasoning rounds with a single deterministic program execution and significantly reducing token usage and latency.  

\begin{figure}
    \centerline{\includegraphics[width=1\linewidth]{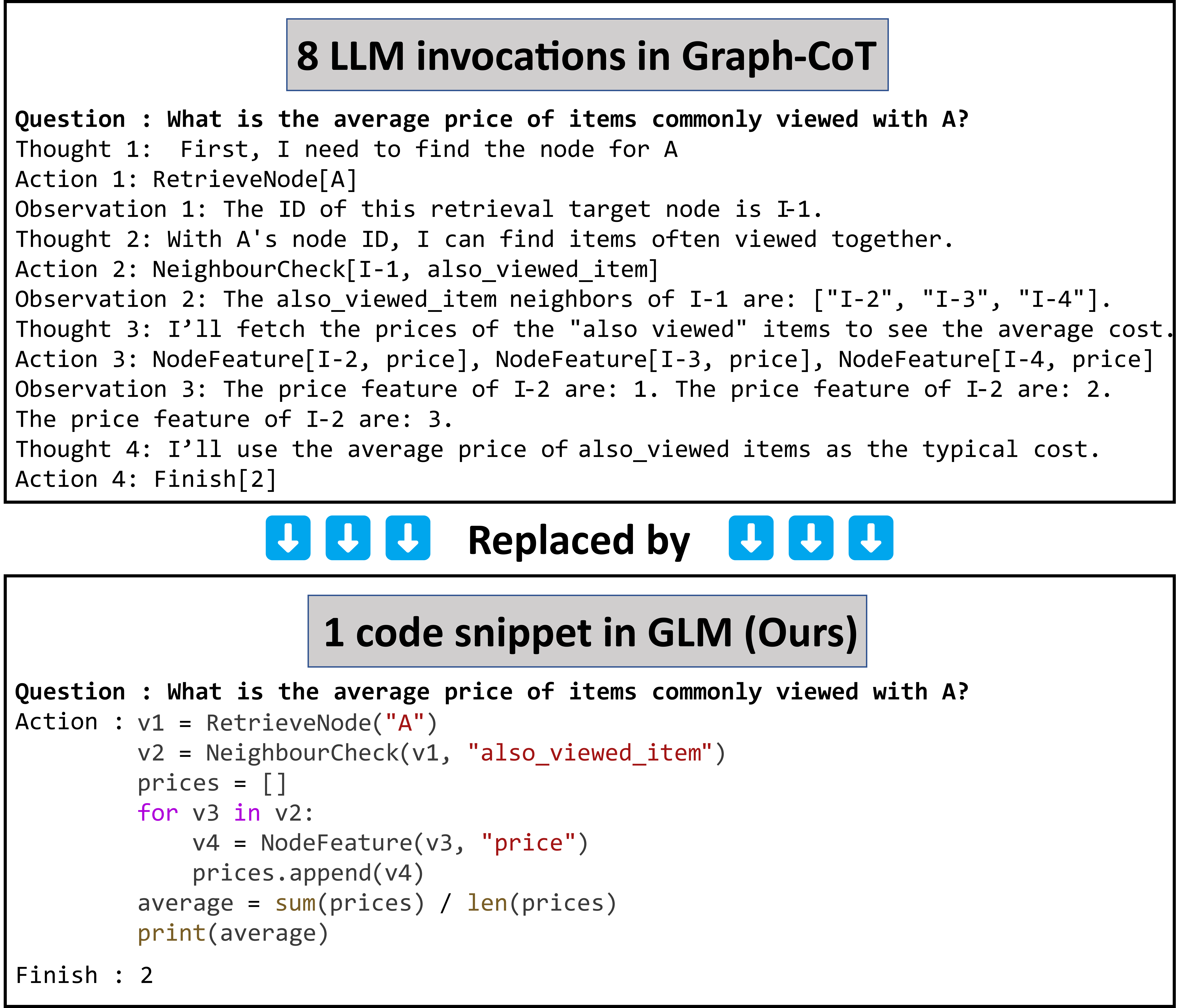}}
    \caption{Code snippet: reducing reasoning steps.}
    \label{fig:example_comparison}
\end{figure}

\textbf{Case Study.}  
As illustrated in Figure~\ref{fig:example_comparison}, for queries requiring information from multiple vertices, a single code snippet can invoke several graph functions within one reasoning round, thereby reducing interaction steps. For queries involving intermediate computations, the snippet performs local data processing (e.g., averaging or intersection) internally, eliminating redundant intermediate storage and minimizing reasoning overhead. This optimization effectively reduces both the number of reasoning rounds and the total token consumption per query.

\section{Graph-CoT-Aware LLM Inference} \label{sec: optimization}

Building on \projectname’s multi-agent Graph-CoT reasoning framework, this section presents graph-cot aware LLM inference optimizations aimed at enhancing end-to-end inference efficiency, reducing end-to-end latency, and improving throughput. Specifically:
(1) We propose a \textit{vertex-centric KV cache reuse model} that leverages the graph structure to share prefix KV caches across queries, reducing latency during the LLM prefill stage;
(2) We introduce an efficient \textit{KV cache scheduling strategy} that minimizes redundant KV computations and improves cache hit rates under high-throughput workloads; 
(3) We design a novel \textit{pipelined execution strategy} that overlaps retrieval with the LLM inference stage, effectively hiding retrieval latency and improving overall system responsiveness.

\subsection{Vertex-Centric KV Cache Reuse Model} \label{sec:vertex_centric_kv}



Unlike text-based RAG systems, which benefit from well-defined chunk boundaries that enable efficient precomputation and prefix KV cache reuse, Graph RAG lacks such structural regularity. In Graph-CoT, retrieval typically occurs at the vertex level, where each vertex contributes a lightweight, atomic fact, such as a feature value or a local structural property (e.g., node degree). As a result, the retrieved context forms a flat and unordered collection of individual facts, limiting the effectiveness of KV cache reuse: during prefill, the model can often reuse only the first retrieved fact, reducing the benefits of caching.  

To address this limitation, we introduce a \textbf{fine-grained retrieval unit} that aggregates richer contextual information per retrieval. Specifically, we define a \emph{vertex chunk} as a reusable, localized subgraph that consists of a central node and its 1-hop neighbours. This structure provides more meaningful context per retrieval, enabling higher KV cache reuse and reducing redundant computation. Formally, the vertex chunk for a given nodeID is defined as follows:

\vspace{-0.5\baselineskip}
\begin{verbatim}
[Node:<nodeID> {k1:v1, k2:v2,...}]
[neighbours:(n1 {k1:v1,...}),(n2 {k1:v1,...}),...]
\end{verbatim}
\vspace{-0.5\baselineskip}

Here, \texttt{<nodeID>} denotes the central node, and each neighbour is represented by a unique identifier (e.g., \texttt{n1}, \texttt{n2}, etc.). The \texttt{k:v} pairs correspond to key-value attributes associated with each node.  

To prevent an excessive number of neighbours from inflating the vertex chunk, we sort each vertex’s neighbours by weight (e.g., node degree) and retain only the top-$k$ nodes with the highest weights.  
Additionally, using a coarser-grained retrieval unit can further reduce reasoning rounds: the LLM can directly select an entire vertex chunk for a given node in a single step rather than iteratively indexing all of the node’s information.  
This approach reduces both the number of reasoning steps and the overall token overhead, improving efficiency in Graph-CoT workflows.

\begin{figure}[t]
    \centerline{\includegraphics[width=1\linewidth]{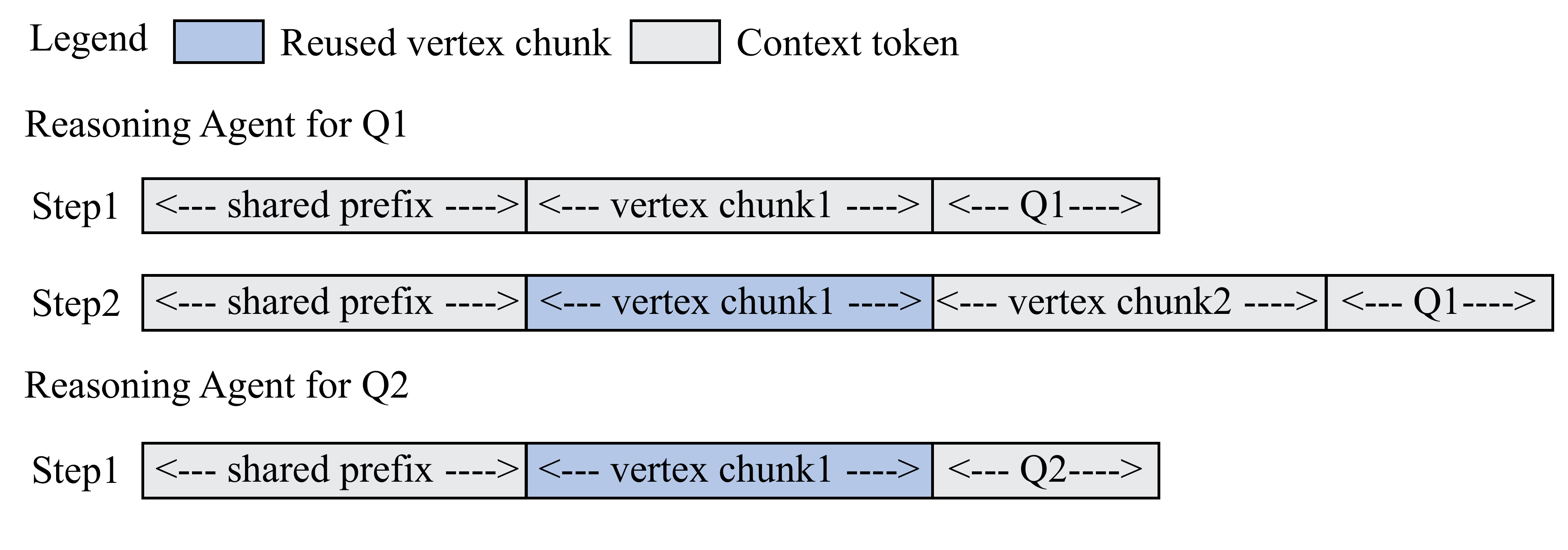}}
    \caption{Example of prefix KV cache reuse model.}
    \label{fig:reuse_model}
\end{figure}

\textbf{Vertex-Centric KV Cache Reuse Model.}  
Consider the reasoning process for a non-deterministic query, where a notebook accumulates information from the Graph RAG over multiple iterations. As illustrated in Figure~\ref{fig:reuse_model}, in the first iteration, the notebook retrieves a vertex chunk associated with a specific node. The reasoning agent processes this chunk (vertex chunk 1), generating the corresponding KV cache. 
Because the notebook retains this information, the same vertex chunk remains relevant across subsequent iterations, allowing its KV cache to be reused, thus eliminating redundant computation. In the second iteration, the same vertex chunk is reused, with its precomputed KV cache directly applied, avoiding recomputation. The reused chunk is highlighted in blue.
Consider now a second, independent query that also retrieves the same vertex chunk. Since the KV cache for this chunk has already been computed, it can be reused without recomputation.

\subsection{Priority-based KV Cache Eviction Policy}

Existing LLM serving systems commonly employ prefix caching to accelerate inference and typically rely on a Least Recently Used (LRU) policy to manage full KV caches. Prefix caching allows queries that share the same prefix to reuse the KV cache directly, avoiding redundant computation. However, LRU policies often evict entries that will be needed in the near future, leading to unnecessary recomputation and inefficiencies. In particular, KV entries corresponding to information retrieved from the graph may be evicted under high workloads, even though they will be reused in subsequent reasoning steps. Conversely, suffixes generated by recently executed agents often change rapidly with task progress and are unlikely to be reused; yet, they are still temporarily retained in the cache.  

To address these limitations, we observe that \textit{the likelihood of future reuse varies across different segments of agents’ inputs}. KV cache entries associated with low-reuse segments should be evicted first, while those with high expected reuse should be retained longer. Based on this insight, we propose a \textbf{priority-based KV cache eviction policy} that accounts for task-specific reuse patterns and manages cache retention more intelligently, reducing redundant computation and improving overall inference efficiency.

Specifically, we introduce a four-tier priority-based KV cache eviction policy to manage prefix KV cache entries based on their expected reuse. Each computed KV entry is assigned a priority level that guides cache retention and eviction decisions under GPU memory constraints. As illustrated in Figure~\ref{fig:agent_prompt}, \projectname employs three LLM-based agents, each driven by a distinct prompt template. Each template is composed of a shared prefix and a query-specific suffix. The shared prefix remains constant across interactions and is reused frequently, making it critical for performance. As such, we assign these entries \textbf{Priority I} and keep them permanently cached throughout execution. For the \textit{reasoning agent}, the \textit{notebook} evolves incrementally during multi-step reasoning, with new content appended at each round. This leads to repeated reuse of notebook-related KV entries within a single query session, justifying their classification as \textbf{Priority II}. These entries are retained for the duration of the session. After the question is resolved, the notebook content is downgraded to \textbf{Priority III}, as it may still be relevant to future queries, particularly when similar reasoning chains or memory contexts are involved. In contrast, components that are less frequently reused, e.g., \textit{missed information} from the action agent or the \textit{question} input in the classification agent, are inexpensive to regenerate. These are assigned \textbf{Priority IV} and are the first to be evicted under memory pressure.

Based on the assigned priorities, we implement an eviction policy that removes KV cache blocks in descending order of evictability when the cache is full. 
We evict KV-cache entries in descending priority: \textbf{Priority IV}, then \textbf{Priority III}, then \textbf{Priority II}. Eviction order is determined by LRU in any priority.
Compared to vLLM’s single-queue LRU, our priority-based KV-cache eviction maintains multiple queues and evicts the lowest-priority entries first. This adds only modest scheduling overhead, which Figure~\ref{fig:optimize_comparison} shows is far outweighed by the resulting throughput and latency gains.

\begin{figure}[t]
    \centerline{\includegraphics[width=1\linewidth]{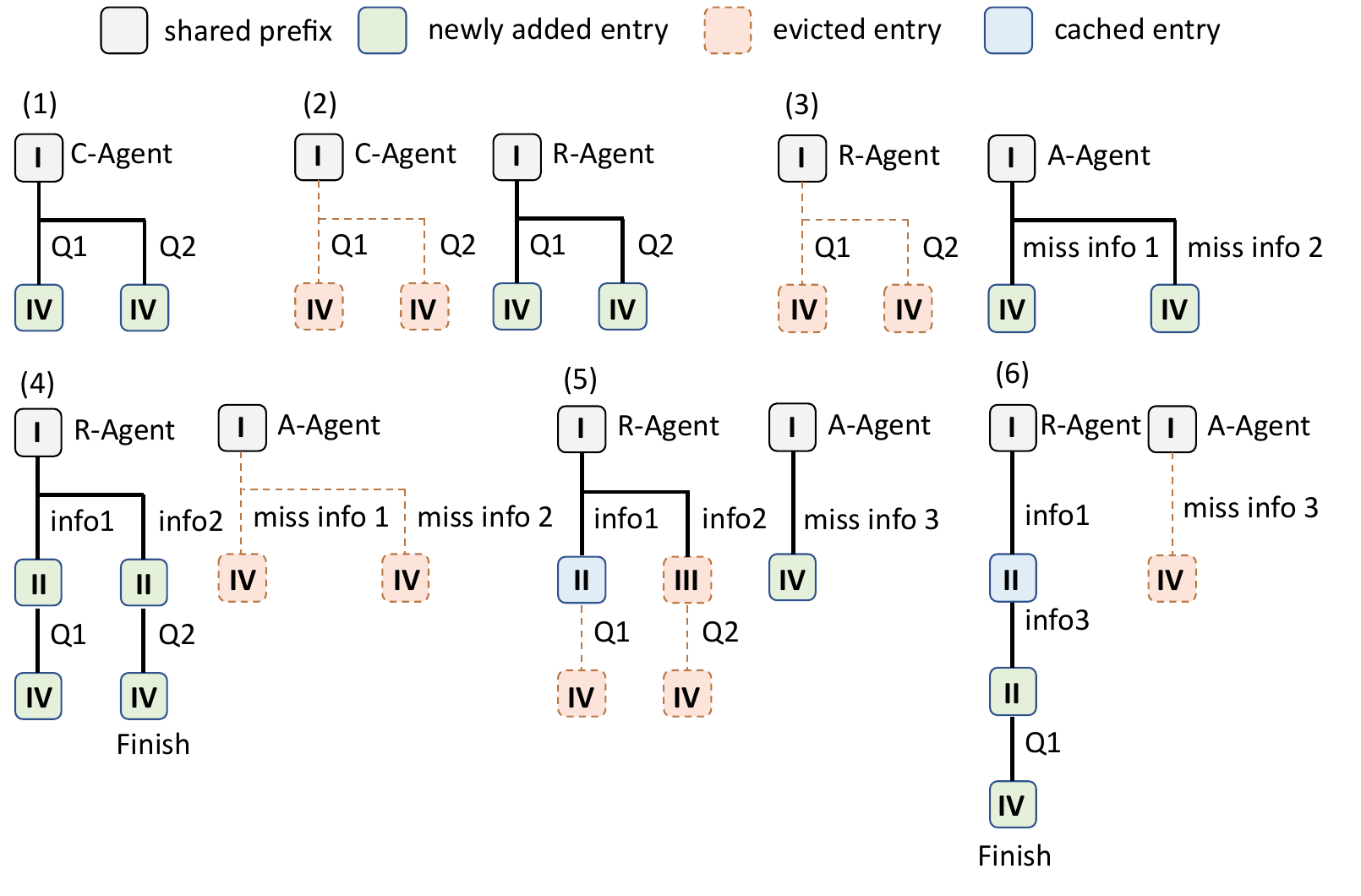}}
    \caption{KV cache scheduling example.}
    \label{fig:kv_cache}
\end{figure}

\textbf{Case Study.}
Figure~\ref{fig:kv_cache} illustrates \projectname with node colors showing KV status:
\textcolor{gray}{gray} = shared prefixes (always cached),
\textcolor{green}{green} = newly added,
\textcolor{blue}{blue} = active and still cached,
\textcolor{red}{red} = evicted.
Two questions (Q1, Q2) proceed as follows:
\begin{enumerate}[leftmargin=*,  nosep]
  \item Shared prefixes (Priority I) are cached. The C-Agent handles Q1 and Q2 (Priority IV) and classifies both as non-deterministic.
  \item The R-Agent receives Q1 and Q2. Notebooks are empty, so both are inserted under the R-Agent prefix; the previous C-Agent KV entries are evicted.
  \item The A-Agent receives the missing-information requests from the R-Agent; earlier R-Agent KV entries are evicted.
  \item After executing the A-Agent snippets, facts \textit{info1} (Q1) and \textit{info2} (Q2) are added to notebooks (Priority II). Q2 is now complete, and \textit{info2} is downgraded to Priority III.
  \item Only one new A-Agent request arrives for \textit{info3} (Q1, Priority II). \textit{info1} remains cached.
  \item The R-Agent finalizes Q1 using the notebook with \textit{info1} and \textit{info3}. Since \textit{info1} is cached, only \textit{info3} and Q1 are reinserted. The response includes \texttt{Finish}, indicating completion.
\end{enumerate}

\subsection{Pipelined Execution Strategy}

\begin{figure}[t]
    \centerline{\includegraphics[width=.9\linewidth]{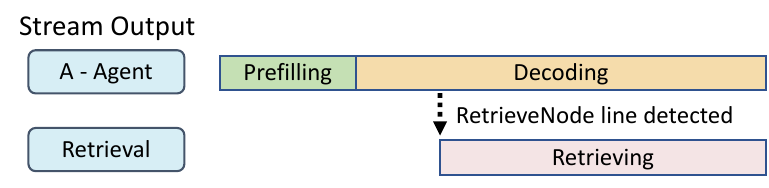}}
    \caption{Pipelined execution strategy.}
    \label{fig:pipeline}
\end{figure}

Given the structure of our multi-agent Graph-CoT framework, we design a pipelining strategy to effectively hide the latency of the Graph RAG retriever. In particular, we observe that when executing code snippets generated by the action agent, the majority of retrieval latency stems from the \texttt{RetrieveNode} function due to the computational complexity involved in performing similarity searches within the vector database. This function typically appears only once, often as the first line of the code snippet. Based on this observation, we divide the action agent’s LLM inference into two stages: (1) the prefill stage and the decoding stage of the line containing the \texttt{RetrieveNode} call, and (2) the decoding stage of the remaining tokens. As illustrated in Figure~\ref{fig:pipeline}, the key idea is to pipeline the execution of the retriever with the second phase of the action agent. The green segment represents the prefill phase of the LLM. The yellow segment corresponds to the initial decoding stage, which includes the decoding of the \texttt{RetrieveNode} function call. Once this call is decoded, the retrieval process is immediately triggered. While the retriever executes (indicated by the pink segment), the LLM concurrently continues decoding the remaining tokens. The dashed arrow illustrates the overlap between retrieval execution and token generation. This overlap is the foundation of our pipelining strategy, which effectively hides retrieval latency and improves overall throughput in multi-agent execution.

To this mechanism, the action agent generates output in a streaming fashion. Once a line containing the \texttt{RetrieveNode} function is detected, the system immediately spawns a pipeline task to execute the retrieval. The retrieved result then replaces the original function call within the partial code sequence, enabling the LLM to complete the remaining decoding based on the updated context.

To further reduce retrieval latency, we integrate a caching mechanism tightly into the pipeline execution. Specifically, we maintain a bounded LRU global cache that maps \texttt{RetrieveNode} inputs (i.e., text queries) to their corresponding output NodeIDs. This cache prevents redundant computation for repeated retrievals. When a \texttt{RetrieveNode} call is detected during the first stage, the system first checks the cache for a match. If a cached result is found, it is reused directly; otherwise, the retrieval is executed and the result is inserted into the cache for future use.

\begin{table}[t]
\centering 
\caption{Summary of GRbench dataset statistics.}
\label{tab:tb-data}
\small
\begin{tabular}{@{}cccccc@{}}  
\toprule
\multirow{2}{*}{Domain} & \multirow{2}{*}{Topic} & \multirow{2}{*}{Abbr} & \multicolumn{2}{c}{Graph Statistics} & \multirow{2}{*}{Questions} \\
\cmidrule(lr){4-5}
 &  &  & \# Nodes & \# Edges &  \\ \midrule
Academic    & DBLP      & ac & $\sim$8M   & $\sim$52M   & 150 \\
E-commerce  & Amazon    & ec & $\sim$9M   & $\sim$313M  & 200 \\
Literature  & Goodreads & li & $\sim$3M   & $\sim$22M   & 240 \\
Healthcare  & Disease   & he & $\sim$47K  & $\sim$4M    & 270 \\
Legal       & Freelaw   & le & $\sim$84M  & $\sim$114M  & 180 \\ \midrule
\textbf{SUM}& -         & -  & \textbf{104M} & \textbf{505M} & \textbf{1040} \\
\bottomrule
\end{tabular}
\end{table}

\begin{table*}[h]
\centering

\caption{Benchmark accuracy comparison of LLM variants evaluated with Rouge-L (R-L) and GPTScore.}
\small
\begin{tabular}{lcccccccccc}
\toprule
\multicolumn{1}{c}{} & \multicolumn{2}{c}{\textbf{Academic}} & \multicolumn{2}{c}{\textbf{E-commerce}} & \multicolumn{2}{c}{\textbf{Literature}} & \multicolumn{2}{c}{\textbf{Healthcare}} & \multicolumn{2}{c}{\textbf{Legal}} \\
\multicolumn{1}{c}{} & \textbf{R-L} & \textbf{GPTScore} & \textbf{R-L} & \textbf{GPTScore} & \textbf{R-L} & \textbf{GPTScore} & \textbf{R-L} & \textbf{GPTScore} & \textbf{R-L} & \textbf{GPTScore} \\ 
\midrule
Base & 0.09 & 0.15 & 0.14 & 0.20 & 0.11 & 0.23 & 0.06 & 0.20 & 0.16 & 0.26     \\
\midrule
Text RAG & 0.09 & 0.14 & 0.25	& 0.31  & 0.15 & 0.28 & 0.04 & 0.20 & 0.21 & 0.26     \\
\midrule
Graph RAG & 0.28 &	0.33 & 0.34 & 0.37 & 0.21 & 0.32 & 0.11 & 0.16 & 0.21 & 0.24     \\
\midrule
Graph-CoT & 0.31 &	0.4 & 0.39 	& 0.44  & 0.42 & 0.53 & 0.33 & 0.46 & 0.46 & 0.53 \\
\midrule
\projectname  &  \textbf{0.55} & \textbf{0.55} & \textbf{0.77} & \textbf{0.79} & \textbf{0.65} & \textbf{0.68} & \textbf{0.62} & \textbf{0.70} & \textbf{0.63} & \textbf{0.64}\\
\bottomrule
\end{tabular}
\label{tab:results}
\end{table*}

\begin{figure*}[ht]
    \centering
    \includegraphics[width=0.88\linewidth]{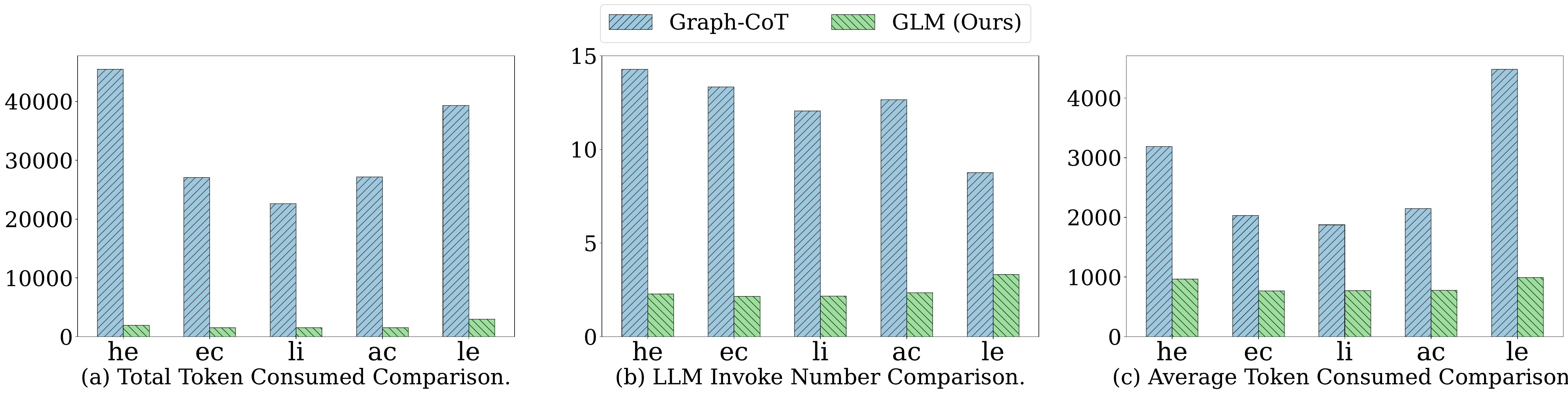}
    \caption{Token consumption comparison.}
    \label{fig:token_consumed_comparison}
\end{figure*}

\begin{figure*}[ht]
    \centering
    \includegraphics[width=0.88\linewidth]{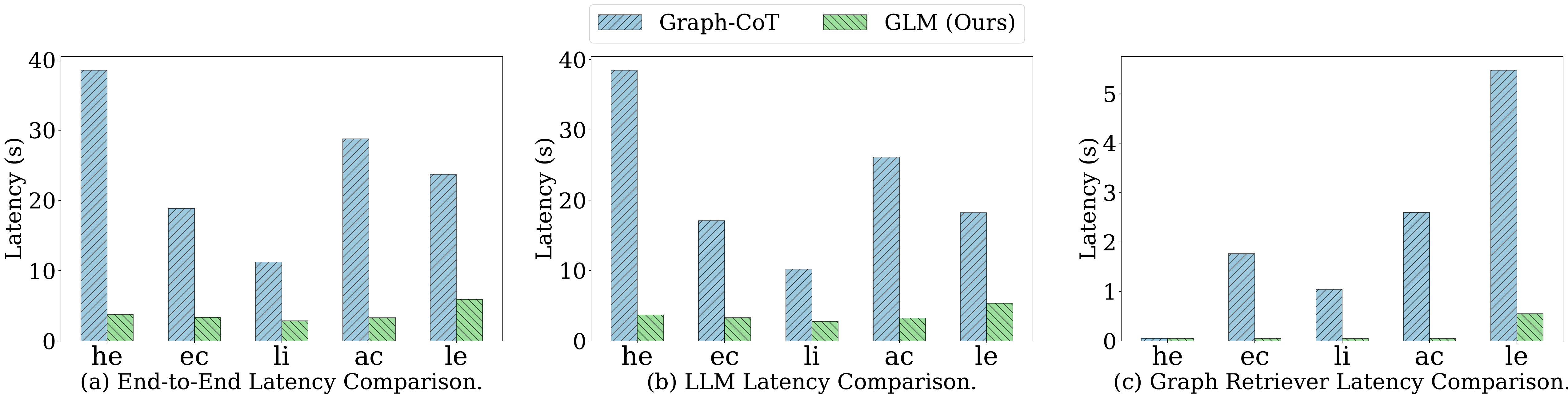}
    \caption{End-to-end latency comparison.}
    \label{fig:end_to_end_latency_comparison}
\end{figure*}

\section{Evaluation}\label{sec:evaluation}

\subsection{Implementation Details}

We implemented a prototype of \projectname in approximately \texttt{3K} lines of Python. It is built on vLLM~\cite{kwon2023efficient} (v0.8.5), with essential modifications to support our Graph-CoT–aware mechanisms.

\stitle{System Architecture.} 
\projectname adopts a modular architecture with two main components: (1) an LLM-based agent module and (2) a Graph RAG retriever module, each implemented as an independent class. The agent module supports specialized agents (\textit{classification}, \textit{reasoning}, and \textit{action}) through customizable prompt templates stored in a \texttt{custom} directory, enabling easy modification and extension to new datasets. The execution follows Algorithm~\ref{alg:workflow}, allowing flexible integration of new agents and control flows.

\stitle{Parallel Batch Processing.} 
For efficient concurrent query handling, \projectname uses a thread pool in which each query runs on an independent thread that shares the same agent and retriever instances to reduce memory overhead. The vLLM backend dynamically batches requests and schedules inference across multiple GPUs, ensuring scalability and high throughput.

\stitle{Fault Tolerance.} 
\projectname ensures robustness through two recovery layers. For code execution errors, the system logs the faulty code and the error message, feeding them back to the LLM for self-correction. For request-level failures, a timeout and recovery mechanism restarts or resumes execution from the last successful step using cached KV pairs, preventing redundant computation and improving reliability under heavy load.

\subsection{Experimental Setup}

\stitle{Hardware.} All experiments are conducted using NVIDIA A100 GPUs with Python 3.8 and Hugging Face Transformers v4.36.2. For retrieval, we use MPNet-v2 as the encoder and implement indexing with FAISS~\cite{johnson2019billion}. The backbone LLM for our main experiments is Qwen3-235B-A22B~\cite{qwen}, with the decoding temperature fixed at $t = 0$ to ensure deterministic outputs. Specifically, we deploy the Qwen3-235B-A22B variant using vLLM v0.8.5 on a cluster of eight NVIDIA A800 SXM4 GPUs, each equipped with 80\,GB of high-bandwidth memory (HBM) and interconnected via NVLINK.

\stitle{Baselines.} We compare \projectname against four widely used categories of baseline methods: standard LLMs (Base LLMs), text retrieval-augmented LLMs (Text RAG LLMs), graph retrieval-augmented LLMs (Graph RAG LLMs), and multi-step reasoning with graph interaction (\textsc{Graph-CoT}):
\begin{itemize}[leftmargin=*, wide, itemindent=0pt, nosep]
    \item \textit{\underline{Base LLMs}:} This baseline tests whether LLMs\cite{qwen} can answer questions using only their internal knowledge, without any access to external data. We provide direct instructions along with the question and allow the LLM to generate an answer.

    \item \textit{\underline{Text RAG LLMs}:} Following \cite{gao2023retrieval}, this method treats the external graph as a plain text corpus. A retriever selects relevant textual content, which is then appended to the prompt as contextual information to guide the LLM's response.

    \item \textit{\underline{Graph RAG LLMs}:} Extending the Text RAG approach, this method linearizes both the retrieved node and its associated subgraph (e.g., 1-hop ego-graph) into a textual sequence, which is then used as input to the Graph RAG LLM~\cite{ye2023language}. This provides a more structured relational context for the model.

    \item \textit{\underline{Graph-CoT}:} As proposed in \cite{jin2024graph}, this method enables LLMs to interact with graphs in an iterative reasoning process. At each step, the model formulates a sub-query, retrieves the necessary information from the graph, and incorporates it into the reasoning trajectory for multi-hop question answering.
\end{itemize}

\stitle{Evaluation Metrics.} We use rule-based and model-based metrics for evaluation. For rule-based assessment, we adopt ROUGE-L (R-L)~\cite{lin2004rouge} to measure surface-level overlap with ground-truth answers. For model-based evaluation, following~\cite{jin2024graph}, we use GPT-4 as a judge to evaluate semantic correctness by comparing the generated answers with the ground truth. The percentage of outputs judged as semantically correct is reported as \textit{GPTScore}~\cite{fu2023gptscore}. We also measure system-level performance using end-to-end latency, throughput (requests per second), and total token consumption to assess efficiency and cost under realistic serving conditions.

\stitle{Datasets.} We evaluate \projectname on GRBench~\cite{jin2024graph}, a benchmark for assessing LLM reasoning over structured knowledge. As shown in Table~\ref{tab:tb-data}, GRBench includes 1,740 manually curated QA pairs from ten real-world graphs across five domains: academia, e-commerce, literature, healthcare, and law. Each question requires one of three reasoning types: factual retrieval, multi-hop reasoning, or inductive inference, providing diverse graph reasoning challenges.

\begin{figure}[t]
    \centering
    \includegraphics[width=0.95\linewidth]{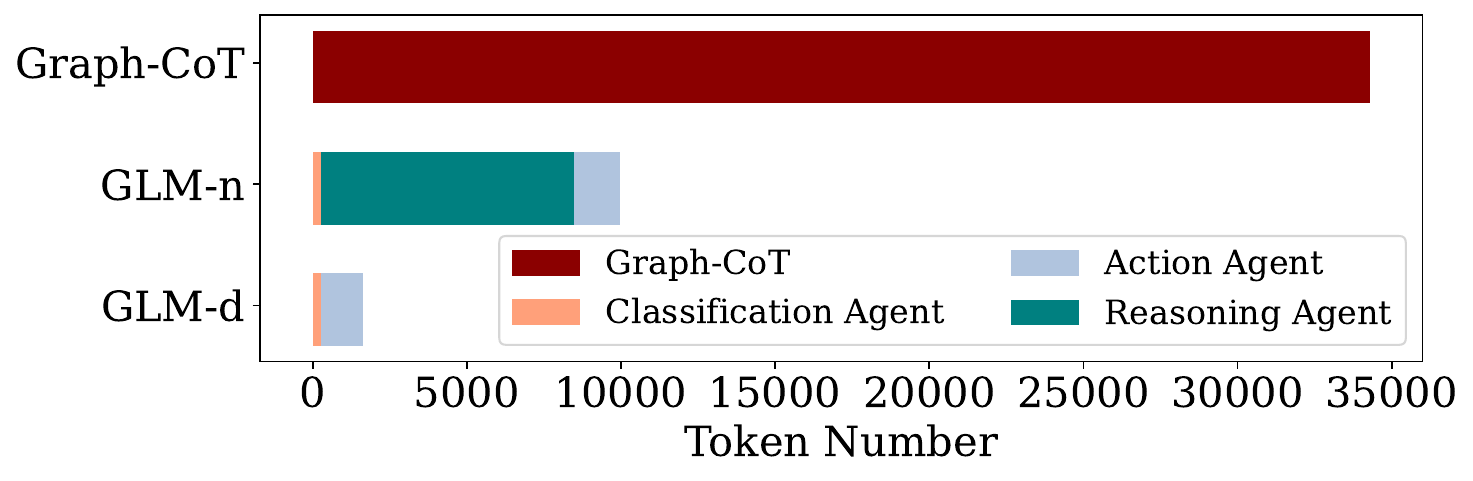}
    \caption{Per-stage token usage comparison.}
    \label{fig:token_consumed_breakdown}
\end{figure}

\subsection{Accuracy Comparison}

We first evaluate the accuracy improvements of \projectname compared to all baseline methods, using both ROUGE-L (R-L) and GPTScore metrics, as reported in Table~\ref{tab:results}. The results demonstrate that \projectname consistently and significantly outperforms all baselines across the evaluated benchmarks. Specifically, \projectname achieves an accuracy gain of 60\% over Base LLMs, 62\% over Text RAG, 55\% over Graph RAG, and 38\% over \textsc{Graph-CoT}. These gains are primarily attributed to \projectname’s novel multi-agent design and code-based reasoning strategy.

A substantial portion of the improvement over \textsc{Graph-CoT} stems from questions that require multi-hop reasoning and information aggregation across multiple nodes. While \textsc{Graph-CoT} often struggles to maintain coherent reasoning over long chains and frequently exceeds the step limit in complex scenarios, \projectname delegates sub-tasks to specialized agents and replaces repetitive LLM reasoning with deterministic, executable code snippets. This results in more structured, accurate, and scalable reasoning.

When compared with Base LLMs, \projectname shows the largest accuracy gain, which depend solely on internal knowledge and often hallucinate or omit details. Text RAG performs better than Base LLMs due to the inclusion of external textual context; however, its retrieval process often fails to recall all relevant information needed for complete reasoning, leading to suboptimal accuracy. Graph RAG LLMs and \textsc{Graph-CoT} perform better than the previous two by incorporating structured graph-aware context and enabling multi-step reasoning. Nonetheless, they still fall short of \projectname due to limitations in reasoning coherence and system efficiency.

In conclusion, \projectname achieves the highest accuracy among all evaluated methods by leveraging its multi-agent Graph-CoT framework, which enables task decomposition, effective use of external graph knowledge, and efficient reasoning through code execution.

\begin{figure}[t]
    \centering
    \includegraphics[width=0.95\linewidth]{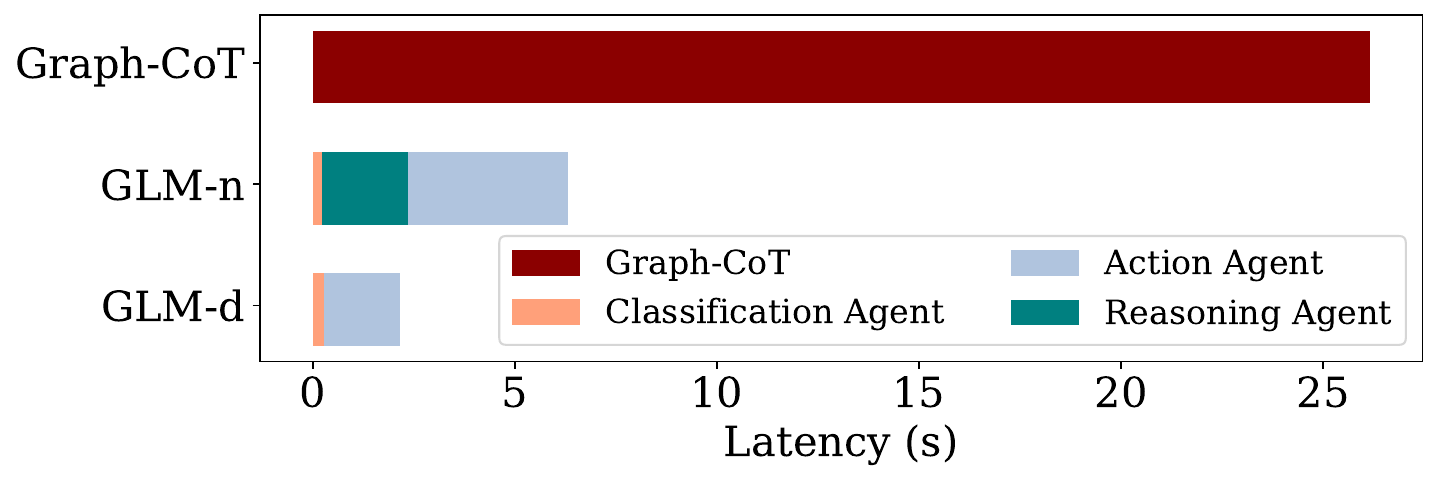}
    \caption{Per-stage latency breakdown comparison.}
    \label{fig:latency_breakdown}
\end{figure}

\subsection{Token Consumption Comparison}\label{sec:token_comparison}

To demonstrate the token efficiency of our multi-agent Graph-CoT reasoning framework, we compare the token consumption of the baseline \textsc{Graph-CoT} with our proposed approach, \projectname.

We first examine the total number of tokens consumed during inference. As shown in Figure~\ref{fig:token_consumed_comparison}, \textsc{Graph-CoT} exhibits substantial token overhead due to iterative multi-step reasoning, long shared prefixes, and redundant context retention. In contrast, \projectname achieves significantly lower token usage. Specifically, the average token consumption per instance ranges from 1,538 $\sim$ 2,974 tokens in \projectname, compared to 22,613 $\sim$ 45,490 tokens in \textsc{Graph-CoT}, representing up to a 95.7\% reduction.

To understand the source of these improvements, we further break down the token usage by measuring (i) the average number of LLM invocations per query and (ii) the average number of tokens consumed per invocation. \projectname requires only 2 $\sim$ 3 LLM calls per instance on average, with each call using approximately 769 $\sim$ 991 tokens. In comparison, \textsc{Graph-CoT} performs an average of 9 $\sim$ 14 invocations per query, each consuming 1,875 $\sim$ 4,483 tokens. These results confirm that \projectname achieves substantial token savings by reducing both the frequency and per-call cost of LLM usage, leading to significantly more efficient inference.

This efficiency stems from three key advantages of \projectname. First, its multi-agent architecture comprising specialized classification, thought, and action agents, ensures that each component operates with minimal, task-specific context, thereby avoiding redundant prompt content. Second, unlike \textsc{Graph-CoT}, which carries the full intermediate reasoning state across steps, \projectname maintains only essential graph-derived facts in a lightweight \textit{notebook}, significantly reducing prompt length. Third, \projectname replaces verbose multi-step reasoning chains with concise, executable code snippets, thereby minimizing the total number of LLM invocations required per task.

We further analyze token composition across agents. As shown in Figure~\ref{fig:latency_breakdown}, for non-deterministic questions, the classification, thought, and action agents contribute to token usage in an approximate ratio of 2.5\% : 82.3\% : 15.1\%. This reflects the fact that reasoning agents process large graph-retrieved vertex chunks, which dominate the token budget. For deterministic questions, token contributions from the classification and action agents are distributed in a ratio of approximately 16.7\% : 83.3\%, where the classification agent consumes substantially fewer tokens, consistent with our architectural analysis in Section~\ref{sec:framework}.

\begin{figure}[t]
    \centering
    \includegraphics[width=0.88\linewidth]{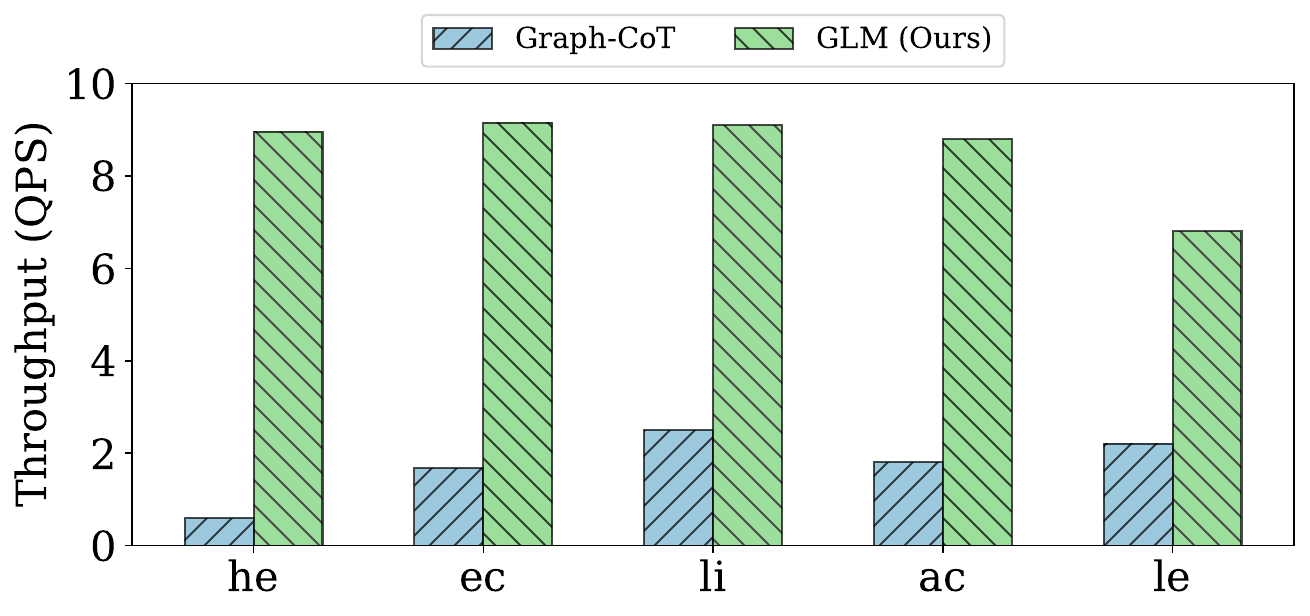}
    \caption{Throughput comparison.}
    \label{fig:throughput_comparison}
\end{figure}


\begin{table*}[h]
\centering
\caption{Fine-grained error taxonomy contrasting Graph-CoT and GLM.}
\small
\begin{tabular}{lcccc}
\toprule
\textbf{Error category} & \textbf{Brief description} & \textbf{Graph-CoT} & \textbf{Graph-CoT with code generation} & \textbf{GLM} \\
\midrule
Unexpected agent output & The agent violates expected format. & 42 (41\%) & 49 (51\%) & 2 (4\%) \\
\midrule
Retrieval process error & The system fetches irrelevant information. & 20 (19\%) & 20 (21\%) & 18 (43\%) \\
\midrule
Code execution error & Generated code fails during runtime. & 2 (2\%) & 20 (21\%) & 20 (48\%) \\
\midrule
Step limit exceeded & The reasoning exceeds the preset step limit. & 38 (37\%) & 8 (8\%) & 2 (4\%) \\
\bottomrule
\end{tabular}
\label{tab:error-comparison}
\end{table*}

\begin{figure*}[t]
    \centering
    \includegraphics[width=.95\linewidth]{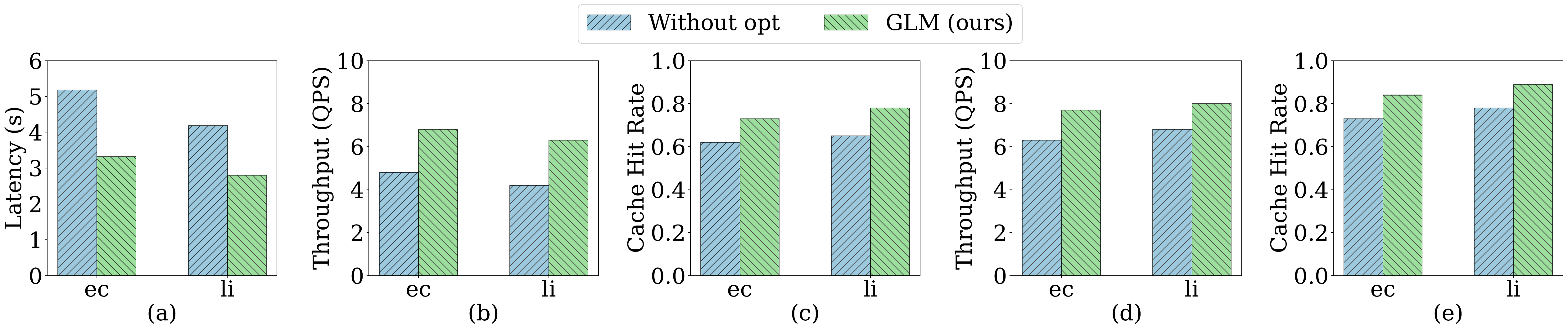}
    \caption{
    Performance breakdown of individual optimizations: \textbf{(a)} evaluates the \textit{pipelined execution strategy}, highlighting latency improvements; \textbf{(b)} and \textbf{(c)} assess the \textit{vertex-centric KV cache reuse model}, reporting gains in throughput and cache hit rate; \textbf{(d)} and \textbf{(e)} analyze the \textit{priority-based KV cache eviction policy}, also in terms of throughput and cache hit rate.
    }    
    \label{fig:optimize_comparison}
\end{figure*}

\subsection{End-to-End Latency Breakdown}

Figure~\ref{fig:end_to_end_latency_comparison} shows the end-to-end latency of \textsc{Graph-CoT} and \projectname. Empirically, \textsc{Graph-CoT} requires 11.3 $\sim$ 38.6 seconds per query, while \projectname completes the same tasks in just 2.8 $\sim$ 5.9 seconds, achieving a latency reduction of up to 74.7\% $\sim$ 90.3\%. \textsc{Graph-CoT} incurs significantly higher latency, primarily due to its single-agent execution model and the lack of system-level optimizations. In contrast, \projectname achieves substantially lower latency through its inherently multi-agent design and three core system optimizations: (1) a pipelined execution strategy that overlaps Graph RAG retrieval with LLM decoding, (2) a vertex-centric KV cache reuse model that efficiently reuses KV cache, and (3) a priority-based KV cache eviction policy that avoids unnecessary computation.

We further analyze end-to-end latency by examining the time distribution across system components. For the Graph RAG retriever, \textsc{Graph-CoT} incurs approximately 0.1 $\sim$ 5.5 seconds for retrieval and 10.1 $\sim$ 38.5 seconds for LLM inference. Retrieval accounts for roughly 0.2\% $\sim$ 23.1\% of the total latency, with the majority of time dominated by multi-step reasoning and LLM decoding. In contrast, \projectname reduces retrieval time to around 0.1 $\sim$ 0.6 seconds and LLM inference time to 2.8 $\sim$ 5.4 seconds. Retrieval constitutes only 1.2\% $\sim$ 9.3\% of the total latency, primarily because it overlaps with LLM decoding, effectively masking most of the retrieval overhead. On smaller datasets, such as Healthcare and Literature, retrieval latency is almost entirely hidden during the decoding phase. However, for larger datasets like Legal, retrieval latency becomes more noticeable due to increased subgraph fetching and preprocessing times that extend beyond the LLM decoding window.

We also examine the latency breakdown of LLM inference within \projectname’s multi-agent framework in Figure~\ref{fig:latency_breakdown}. For non-deterministic questions, the classification agent, reasoning agent, and action agent contribute to latency in a ratio of 3.7\% : 42.1\% : 54.2\%. This distribution reflects a trade-off between input and output token lengths: the reasoning agent typically processes more input tokens, increasing prefill and decoding overhead, while the action agent produces longer output sequences (e.g., code snippets), which require more decoding steps. Both agents benefit from relatively high KV cache hit rates on input tokens, mitigating the incremental cost of processing lengthy contexts. Consequently, their total inference times remain comparable. For deterministic questions, the classification and reasoning agents account for 12.4\% to 87.6\% of the latency, consistent with the token distribution reported in Section~\ref{sec:token_comparison}.



\subsection{Throughput Comparison}\label{sec:throughput_comparison}

Throughput is defined as the number of question–answer pairs processed per unit time. As shown in Figure~\ref{fig:throughput_comparison}, \textsc{Graph-CoT} achieves a modest throughput of only 0.6 $\sim$ 2.2 queries per second. In contrast, our proposed system, \projectname, delivers higher throughput, reaching 6.8 $\sim$ 9.1 queries per second. This improvement stems from two core design principles: (1) a multi-agent architecture that parallelizes and accelerates the reasoning process, and (2) a \textit{vertex-centric KV cache reuse model} combined with an efficient \textit{KV cache eviction policy} that maximizes cache reuse and eliminates redundant computations. Together, these optimizations make \projectname particularly well-suited for high-throughput scenarios under constrained compute budgets, achieving up to 3.2$\times$ to 15.1$\times$ higher throughput than \textsc{Graph-CoT}.

\subsection{Efficiency of Different Optimizations}

We analyze the impact of different system-level optimizations on computational efficiency, focusing on three key components: the \textit{vertex-centric KV cache reuse model}, \textit{priority-based KV cache eviction policy}, and the \textit{pipelined execution strategy}. To evaluate the cache-related optimizations designed primarily for non-deterministic queries, we augment the original datasets with additional non-deterministic questions. In particular, the e-commerce and literature datasets are extended due to the limited number of such queries in the original benchmark. Detailed results are presented in Figure~\ref{fig:optimize_comparison}, covering latency, throughput, and cache hit ratio comparisons.

\noindent\textbf{Efficiency of Pipeline Execution Strategy.}  
We evaluate the pipelined execution strategy, which overlaps the execution of retrieval with LLM decoding. Without pipelining, average response latency ranges from 4.3 $\sim$ 10.2 seconds. With pipelining enabled, latency is reduced to 2.8 $\sim$ 5.3 seconds—an improvement of up to 47.8\%. This gain is achieved by precomputing retrieval results and masking I/O latency during token generation, thereby reducing idle time and improving responsiveness in interactive workloads.

\noindent\textbf{Efficiency of Vertex-Centric KV Cache Reuse Model.}  
This optimization improves the information granularity in Graph RAG retrieval by enabling prefix segments of the KV cache to be shared across different queries. As a result, it reduces redundant computation and improves execution efficiency. Experimental results show a 41.6\% increase in throughput and a 17.7\% boost in cache hit rate. Moreover, coarser-grained retrieval not only enhances reuse opportunities but also shortens the overall reasoning steps required by the GLM, further amplifying the throughput benefit.

\noindent\textbf{Efficiency of the Priority-Based KV Cache Eviction Policy.}  
This mechanism improves cache utilization by assigning eviction priorities based on token frequency and reuse potential. As shown in Figure~\ref{fig:optimize_comparison}, it improves overall throughput by 14.3\% and reduces the cache miss rate by 15.1\%, showing its effectiveness in managing limited GPU memory and reducing redundant recomputation.

In conclusion, \projectname's performance optimizations effectively improve throughput, increase the cache hit rate, and reduce latency across various configurations, demonstrating the overall effectiveness of our system-level design.

\subsection{Error Analysis of Code Generation}

In this section, we analyze failure cases in Graph-CoT and \projectname. Table~\ref{tab:error-comparison} categorizes errors into four types: \emph{unexpected agent output}, \emph{retrieval process error}, \emph{code execution error}, and \emph{step limit exceeded}.  

In Graph-CoT, the most frequent failure (41\%) arises from \emph{unexpected agent output}, caused by the single-agent design handling long contexts, which often leads to the loss of intermediate information. The second most common error (37\%) is \emph{step limit exceeded}, which occurs when queries require information from multiple nodes, resulting in reasoning rounds that exceed the allowed limit.  

Introducing the code generation optimization compresses multi-step reasoning into a single executable snippet, reducing the number of steps per query (e.g., from 8 to 4). This significantly lowers the incidence of \emph{step limit exceeded} errors. However, this optimization increases \emph{code execution errors}, as the LLM now synthesizes full code snippets rather than selecting from a predefined set of graph-interaction functions.  

In \projectname, the multi-agent framework further mitigates \emph{unexpected agent output}, reducing it from 41\% to 2\% because each agent operates within a minimal, task-specific context and is less likely to lose track of intermediate information. Overall, code generation introduces a trade-off: \emph{code execution errors} increase (2\% $\rightarrow$ 20\%) while \emph{step limit exceeded} errors drop dramatically (38\% $\rightarrow$ 8\%). 
Additional optimizations in \projectname further reduce \emph{unexpected agent output}, demonstrating the effectiveness of the multi-agent and inference-aware design.



\section{Discussion}

\stitle{Impact on AI–Data System Co-Design.}
This work bridges LLM reasoning and graph data management by introducing a multi-agent \textit{Graph-CoT} framework co-designed with system-level serving optimizations. The proposed classification–reasoning–action decomposition offers a reusable modular abstraction for accelerating long-memory agents\cite{zhang2025survey}, graph-enhanced LLMs\cite{liu2024knowledge}, and retrieval-augmented systems\cite{peng2024graph} across tasks such as knowledge-graph QA, structured information extraction, and task planning. Moreover, our vertex-centric KV-cache reuse and priority-based eviction policies generalize to multi-agent and structured RAG serving pipelines, providing a practical path toward high-throughput, low-latency inference at scale. We believe this work opens new directions for research at the intersection of data systems and LLMs, advancing scalable, structure-aware reasoning systems.

\stitle{Limitations and Future Work.}  
Graph-CoT opens several promising directions for future research. We highlight two main limitations of the current work and the corresponding opportunities for improvement in future research.  
First, the accuracy of \projectname{} remains bounded by the capability of the underlying backbone LLM. Future research may explore techniques such as knowledge distillation from larger models, lightweight fine-tuning (e.g., adapter or LoRA), tighter retrieval–reasoning calibration, and structure-aware prompting to reduce dependence on raw model capacity. 
Second, the graph retrieval process can introduce significant latency, particularly in the absence of GPU acceleration, and it often constitutes the dominant system bottleneck. Future work could investigate optimizing the retrieval pipeline and overall system architecture to reduce latency and resource consumption while preserving reasoning quality.

\section{Conclusion}

In this work, we revisited Graph-CoT reasoning and addressed the scalability limitations of existing single-agent approaches, which suffer from high token cost, latency, and limited accuracy. We introduced \projectname, a multi-agent Graph-CoT framework co-designed with an optimized LLM serving architecture and system-level optimizations tailored for Graph-CoT workloads. \projectname improves accuracy while dramatically reducing token usage and inference latency. Extensive experiments demonstrate that \projectname delivers substantial accuracy gains with significant efficiency improvements, making structured LLM reasoning practical at scale.


\balance
\bibliographystyle{ACM-Reference-Format}
\bibliography{ref}
\end{document}